\documentclass[10pt,journal,twoside]{IEEEtran}
\usepackage{amsmath,amsfonts}
\usepackage{algorithmic}
\usepackage{array}
\usepackage[caption=false,font=scriptsize,labelfont=sf,textfont=sf]{subfig}
\usepackage{textcomp}
\usepackage{stfloats}
\usepackage{url}
\usepackage{verbatim}
\usepackage{graphicx}
\usepackage{cite}
\usepackage[numbers]{natbib}
\usepackage{booktabs}
\usepackage{multirow}
\usepackage{tabularx}
\usepackage{tablefootnote}
\usepackage{threeparttable}
\usepackage{caption}
\usepackage{pifont}
\usepackage{algorithm}

\begin{document}

\title{Is the Trigger Essential? A Feature-Based Triggerless Backdoor Attack in Vertical Federated Learning}

\author{
    Yige Liu, Yiwei Lou, Che Wang, Yongzhi Cao, \textit{Senior Member, IEEE}, and Hanpin Wang
    \thanks{Yige Liu, Yiwei Lou, Che Wang, Yongzhi Cao, and Hanpin Wang are with the Key Laboratory of High Confidence Software Technologies (Peking University), Ministry of Education; School of Computer Science, Peking University, Beijing, China.~(e-mail: yige.liu@stu.pku.edu.cn; cyfqylyw@gmail.com; wangche02@gmail.com; caoyz@pku.edu.cn; whpxhy@pku.edu.cn)}
    \thanks{Yongzhi Cao is also with the Zhongguancun Laboratory, Beijing, China.}
    \thanks{Corresponding author: Yongzhi Cao}
}

\maketitle

\begin{abstract}
    As a distributed collaborative machine learning paradigm, vertical federated learning~(VFL) allows multiple passive parties with distinct features and one active party with labels to collaboratively train a model. Although it is known for the privacy-preserving capabilities, VFL still faces significant privacy and security threats from backdoor attacks. Existing backdoor attacks typically involve an attacker implanting a trigger into the model during the training phase and executing the attack by adding the trigger to the samples during the inference phase. However, in this paper, we find that triggers are not essential for backdoor attacks in VFL. In light of this, we disclose a new backdoor attack pathway in VFL by introducing a feature-based triggerless backdoor attack. This attack operates under a more stringent security assumption, where the attacker is honest-but-curious rather than malicious during the training phase. It comprises three modules: label inference for the targeted backdoor attack, poison generation with amplification and perturbation mechanisms, and backdoor execution to implement the attack. Extensive experiments on five benchmark datasets demonstrate that our attack outperforms three baseline backdoor attacks by 2 to 50 times while minimally impacting the main task. Even in VFL scenarios with 32 passive parties and only one set of auxiliary data, our attack maintains high performance. Moreover, when confronted with distinct defense strategies, our attack remains largely unaffected and exhibits strong robustness. We hope that the disclosure of this triggerless backdoor attack pathway will encourage the community to revisit security threats in VFL scenarios and inspire researchers to develop more robust and practical defense strategies.
\end{abstract}


\section{Introduction} 

\IEEEPARstart{V}{ertical} federated learning~(VFL) is a collaborative machine learning paradigm that involves multiple passive parties and one active party to train a model. It originates from federated learning~\cite{mcmahan2016federated} and split learning~\cite{vepakomma2018split}, and is juxtaposed with the federated learning classifications of horizontal federated learning~(HFL) and federated transfer learning categorized by~\citet{yang2019federated} based on the different data divisions. Among them, VFL is distinguished by its data distribution among participants: passive parties possess distinct features, while the active party holds the labels. During training, each passive party converts its features into embeddings using the bottom model and sends them to the active party. After receiving embeddings from all passive parties, the active party aggregates them and employs the top model to predict labels. Subsequently, the VFL model training is guided by comparing the predicted labels with the ground-truth labels. Driven by advancements in machine learning and the increasing need for privacy preservation, VFL has been widely studied and applied. For example, in academic research, \citet{qiu2023your}, \citet{he2024backdoor} and \citet{qiu2024hashvfl} study attacks and defenses in VFL, and \citet{gong2024multi} and \citet{yang2024openvfl} aim to propose a secure VFL framework. Furthermore, in the financial sector, WeBank utilizes VFL to implement federated risk control for small and microenterprise loans~\cite{cheng2020federated}.

While VFL can offer privacy preservation through data localization and model separation, it remains vulnerable to some privacy and security threats, including label inference attacks~\cite{fu2022label,liu2024similarity}, feature reconstruction attacks~\cite{vu2023active,qiu2024hashvfl}, and backdoor attacks~\cite{naseri2023badvfl,he2023backdoor,chen2024universal}. Notably, backdoor attacks are typically initiated by a malicious passive party adding a trigger to the sample for misclassifying its prediction during the inference phase. Inspired by backdoor attacks in HFL scenarios, \citet{bai2023villain} adapted this concept of implanting triggers into models for VFL scenarios and conducted the first study of backdoor attacks in VFL. Following their work, numerous existing VFL backdoor attacks primarily focus on generating new triggers to efficiently implement attacks~\cite{naseri2023badvfl,he2023backdoor,chen2024universal}. However, we find that backdoor attacks in VFL do not face the same practical scenarios as backdoor attacks in HFL. In HFL, clients collaboratively train the same model during the training phase and use this model separately during the inference phase. This necessitates attackers to implant a trigger into the model for a backdoor attack. Conversely, in VFL, the distributed ownership of the model across participants requires the involvement of passive parties during the inference phase, allowing an attacker to directly alter the embeddings sent to the active party for backdoor attacks. This distinction prompts us to explore the following question.

\begin{center}
    \em Is the trigger essential in VFL?
\end{center}

To answer this, we propose a feature-based triggerless backdoor attack and disclose a new attack pathway in VFL. This attack leverages the label characterization capability of a well-trained bottom model. Instead of implanting triggers during the training phase, the attacker manipulates model predictions by replacing benign embeddings with malicious ones exclusively during the inference phase. Specifically, our novel attack comprises three modules: label inference, poison generation, and backdoor execution. During the training phase, the attacker simply records the embeddings generated by the bottom model at each epoch, without engaging in any actions that breach the VFL protocol. This ensures that the attack is \textit{honest-but-curious} rather than \textit{malicious}. During the inference phase, the \textit{malicious} attacker first infers sample labels from the recorded embeddings using a clustering algorithm. Next, the attacker calculates the embedding density centers for different labels and generates a set of malicious embeddings as poisons using amplification and perturbation mechanisms. After completing these preparations, the attacker identifies the target samples and replaces their embeddings with malicious ones to execute the backdoor attack. Without implanting the trigger during the training, our attack can be implemented under a more stringent security assumption and is difficult to detect by defenses. Additionally, since our attack depends on the label characterization capability of embeddings, existing research on enhancing the latent representation of passive parties~\cite{huang2023vertical,wu2024practical,rashad2024tabvfl} may further improve our attack performance.

The success of this new attack pathway across extensive experiments demonstrates that triggers are not essential for backdoor attacks in VFL scenarios. Specifically, we conduct evaluations on five benchmark datasets in the multi-party VFL scenario. Compared to three existing baseline backdoor attacks, it achieves a success rate that is 2 to 50 times higher in terms of overall performance. In addition, we evaluate the impact of some potential factors introduced by our attack on the performance. The results demonstrate that our attack maintains high accuracy even in a VFL scenario with 32 passive parties and with only one set of auxiliary data. We also investigate the impact of hyperparameters including learning rate, batch size, training epochs, and models. Furthermore, the attack performance under defense strategies is also taken into account. We first assess it under eight existing defenses across three perspectives: general defense, defense against label inference, and defense against backdoor attacks. The results show that our attack remains largely unaffected and demonstrates strong robustness. Since existing defenses mainly target the training phase, we further propose a novel defense strategy specifically designed for the inference phase and evaluate our attack's performance against it. Notably, even this targeted defense struggles to counter our attack on complex datasets. Moreover, to provide a comprehensive research, we additionally discuss the impact of embedding layer and learning rate adjustment on the performance of our attack.

The main contributions are summarized as follows:

\begin{itemize}
    \item We present the first observation that triggers are not essential in VFL scenarios and disclose a new backdoor attack pathway in VFL by proposing a novel feature-based triggerless backdoor attack. This attack comprises three modules: label inference, poison generation, and backdoor execution, and can be carried out under more stringent security assumptions.
    \item We compare the performance of our attack with three existing baseline attacks on five datasets, evaluate potential influencing factors, investigate the impact of hyperparameters on attack performance, and assess our attack's robustness under four different types of defense strategies. The results demonstrate that our triggerless backdoor attack exhibits strong performance and robustness in multi-party VFL scenarios.
    \item Our disclosure of this simple yet effective triggerless backdoor attack pathway may encourage the community to reassess security in VFL scenarios, as it is easy to implement, challenging to detect, and delivers great value at minimal cost. Additionally, it can inspire researchers to develop more robust and practical defense strategies.
\end{itemize}

\section{Preliminaries}

In this section, we present the VFL formalization and summarize related works on backdoor attacks in VFL.

\subsection{Vertical Federated Learning}
\label{sec:VFL}

As a distributed machine learning paradigm, VFL allows multiple participants to collaboratively train a model. In VFL, the participants share the same user or sample space but hold different feature spaces. Specifically, the passive parties hold distinct raw feature data, while the active party holds the labels. Based on whether the active party possesses additional features, \citet{liu2024vertical} categorized VFL into two types: aggVFL and splitVFL. In aggVFL, the active party holds only labels, while in splitVFL, the active party holds both labels and features. In the paper, we concentrate on aggVFL and formalize its training process as follows.

\begin{figure}[t]
    \centering
	\includegraphics[width=\columnwidth]{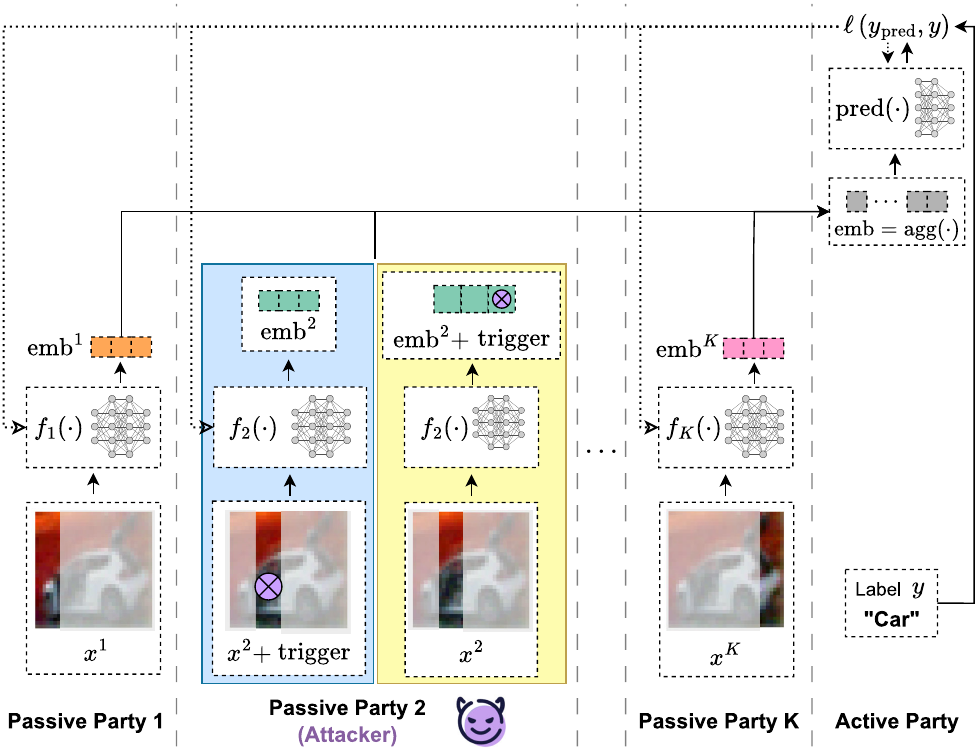}
    \caption{An illustration of the backdoor attack in VFL. During the training phase, the attacker can implant triggers into the model in two ways: by adding the trigger to the raw data~(blue background) or the embeddings~(yellow background). Forward propagation is represented by the solid arrow and backpropagation by the dashed arrow.}
    \label{fig:backdoor}
\end{figure}

As illustrated in Fig.~\ref{fig:backdoor}, the VFL model is collaboratively trained with $N$ samples by $K$ passive parties, each holding distinct features, and one active party holding the labels~\cite{liu2024vertical}. The training process is divided into two phases: forward propagation and backpropagation.

During forward propagation, each passive party $k$ utilizes its local embedding model~(bottom model) $f_k\left(\cdot\right)$ parameterized by $\theta_k$ and its local data $x_i^k$ from datasets $\left\{x_i^k\right\}_{i=1}^N$ to compute embeddings $\mathrm{emb}_i^k=f_k\left(\theta_k;x_i^k\right)$, which are then sent to the active party. The active party first aggregates the embeddings from all passive parties into a single embedding $\mathrm{emb}_i=\mathrm{agg}(\mathrm{emb}_i^1,\dots,\mathrm{emb}_i^K)$ by a certain aggregation algorithm. Subsequently, it trains the prediction model~(top model) $\mathrm{pred}\left(\cdot\right)$ parameterized by $\varphi$ through the aggregated embedding $\{\mathrm{emb}_i\}_{i=1}^N$ and ground-truth labels $\{y_i\}_{i=1}^N$ using the loss function
\begin{equation}
    \mathcal{L}\left(\theta,\varphi;x,y\right)=\frac{1}{N} \sum_{i=1}^N \ell\left(\mathrm{pred}\left(\varphi;\mathrm{emb}_i\right),y_i\right),
\end{equation}
where $\ell\left(\cdot\right)$ denotes the sample loss such as cross-entropy loss or mean squared error loss.

During backpropagation, the active party sends gradients $\nabla_{\varphi}\mathcal{L}$ to update the prediction model and gradients $\nabla_{\mathrm{emb}_i^k}\mathcal{L}=\frac{\partial\mathcal{L}}{\partial\mathrm{emb}_i^k}$ to each passive party. Employing the chain rule, each passive party updates its bottom model with these gradients. The training process iterates until the model converges.

\subsection{Backdoor Attacks}
\label{sec:backdoor}

The backdoor attack aims at making the model get misclassified predictions by implanting backdoor triggers into the model. If these misclassified predictions do not consistently correspond to a specific label, the attack is classified as an \textit{untargeted} backdoor attack. In contrast, if the model consistently misclassifies predictions as a specific label, the attack is identified as a \textit{targeted} backdoor attack.

In VFL scenarios, untargeted backdoor attacks are typically implemented as adversarial attacks. \citet{pang2022attacking} discovered that certain inputs from the passive party may dominate the training of the VFL model, steering its predictions toward the attacker's intended outcome. They first proved the existence of these adversarial dominating inputs~(ADIs) in two-party VFL, and then proposed a gradient-based adversarial sample generation method for untargeted backdoor attacks. Furthermore, \citet{chen2022graph} introduced an adversarial attack on graph neural networks by privacy leakage and the gradients of pairwise nodes, which generates adversarial samples based on the addition of noisy embedding of global nodes.

Since the features and labels in VFL are held by the passive parties and the active party, respectively, a targeted backdoor attack typically consists of two phases: label inference and backdoor implantation. The label inference phase aims to identify samples with the source label, ensuring that the backdoor can be implanted into the correct sample. During the backdoor implantation phase, the attacker implants the backdoor into the VFL model by adding triggers to the raw data or the embedding.

\begin{figure*}[t]
    \centering
    \includegraphics[width=\textwidth]{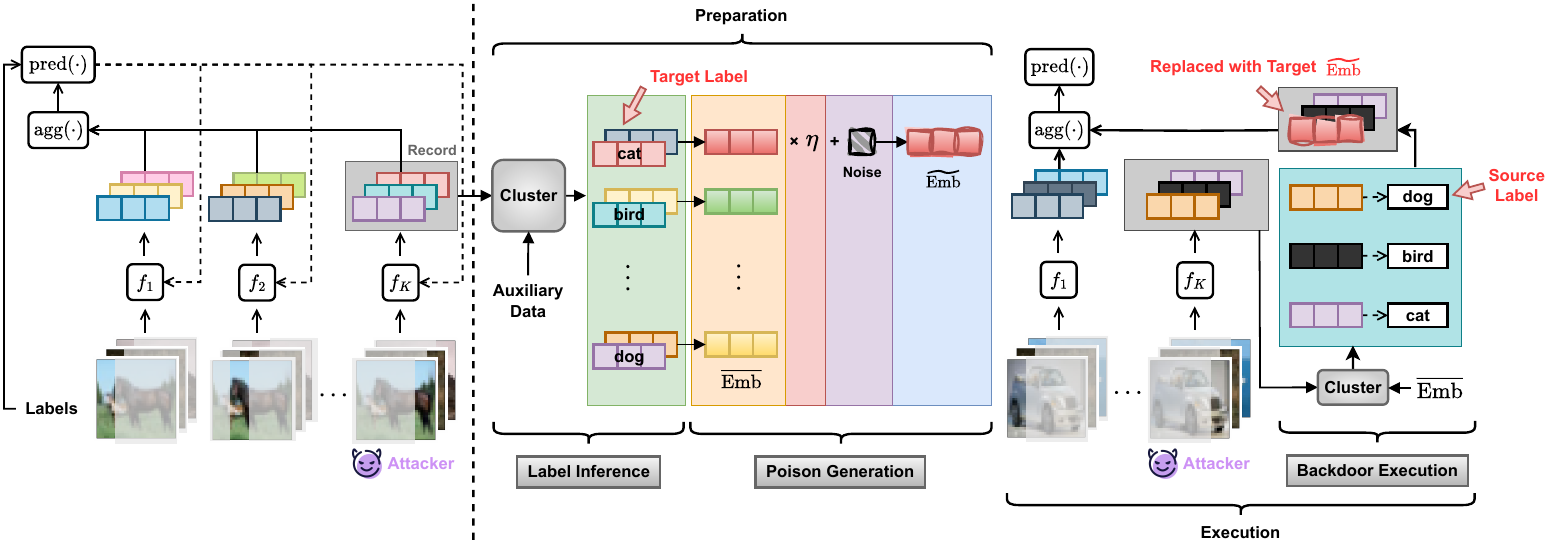}
    \caption{An illustration of the feature-based triggerless backdoor attack in VFL. The attack consists of three modules: label inference, poison generation, and backdoor execution. The first two modules prepare the necessary poison for the attack, while the third module execute the attack. Left is the training phase, and right is the inference phase.}
    \label{fig:triggerless}
\end{figure*}

\citet{bai2023villain} were the first to focus on targeted backdoor attacks in VFL. They proposed a backdoor attack method named VILLAIN and utilized a new trigger inspired by the strip-based backdoor~\cite{barni2019new}, which is a pattern of two positive values followed by two negative values. \citet{naseri2023badvfl} proposed BadVFL, which analyzes the backpropagated gradients to identify the most influential parts of the raw data, subsequently implanting cross triggers into these identified areas. Meanwhile, \citet{he2023backdoor} designed a trigger by calculating the density center of similar embeddings via the mean-shift algorithm, and introduced a scaling factor to enhance the impact of the backdoor vectors on the entire VFL model in backdoor attacks.

Beyond fixed triggers, optimization-based triggers are also used in existing works. \citet{chen2023practical} proposed the TECB method to train the trigger with a minimal set of labeled samples. They addressed the issue of trigger differences between the training and inference phases by employing gradient replacement to refine the model. Inspired by the model completion technique~\cite{fu2022label}, \citet{gu2023lr} developed the LR-BA method, which generates triggers through the reconstructed surrogate model, ensuring the successful activation of triggers in backdoor tasks. Moreover, \citet{chen2024universal} introduced a universal backdoor attack method for binary classification tasks. It divides the training task into two parts, the model task and the backdoor task, facilitating the acquisition of an effective trigger during the backdoor implantation phase.

\section{Methodology}

In this section, we elaborate on our proposed feature-based triggerless backdoor attack. As shown in Fig.~\ref{fig:triggerless}, in our attack, the attacker only records the embeddings generated by the bottom model during the training phase without violating any protocol, while the backdoor attack is launched exclusively during the inference phase. This attack comprises three main modules: label inference, poison generation, and backdoor execution. The first two modules prepare the backdoor poison necessary for the attack, while the third module executes the attack. Specifically, during the inference phase, the attacker first infers the sample labels and records the embeddings corresponding to different labels using the label inference module. Next, the recorded embeddings are utilized to generate a valid set of malicious embeddings in the poison generation module. Finally, the attacker identifies the target samples and replaces their embeddings with the malicious embeddings to execute the backdoor attack. The full algorithm is provided in Appendix~A.

\subsection{Threat Model}

We define the threat model for our triggerless backdoor attack in VFL, which includes the attacker's goals, knowledge and capabilities. Since our attack does not require the implantation of triggers into the model, without loss of generality, we assume that all passive parties train the model following the protocol during the training phase. However, they are \textit{honest-but-curious} to collect and analyze any data they can legitimately obtain. During the inference phase, any passive party may launch a backdoor attack as a \textit{malicious} attacker, but they do not collude with each other to perform a collusive attack. Unlike traditional backdoor attacks, where the attacker is malicious in both the training and inference phases, our proposed backdoor attack operates under a more stringent security assumption: the attacker is \textit{honest-but-curious} during the training phase and \textit{malicious} only during the inference phase.

\subsubsection{Attacker's Goal} The attacker aims to implement the targeted backdoor attack. Similar to traditional trigger-based targeted backdoor attacks, our proposed triggerless targeted backdoor attack aims to achieve the following: during the inference phase, sending the original benign embeddings of samples from any label category to the top model yields the correct label, while sending malicious embeddings causes the top model to incorrectly predict the target label $l_t$.

\subsubsection{Attacker's Knowledge} The attacker possesses a local dataset along with immutable indexes. During the training phase, the attacker adheres to the protocol for participating in the VFL model training, gaining all details related to the bottom model $f_a\left(\cdot\right)$, such as parameters and output embeddings for each epoch. Additionally, the attacker can obtain gradients backpropagated by the active party. The attacker also has a limited small quantity of labeled auxiliary data $Aux=\{X^u,Y^u\}$, which can be acquired through purchase or is publicly accessible. This assumption is consistent with those made in existing related works~\cite{naseri2023badvfl,he2023backdoor,bai2023villain,chen2023practical}. Given that all parties involved in VFL collaborate on training objectives and typically share a small amount of training data examples before training, it is realistic and reasonable for attackers to obtain limited labeled auxiliary data. However, the attacker lacks any information regarding the active party, including its model structures, parameters and labels.

\subsubsection{Attacker's Capability} During the training phase, the attacker trains a bottom model following the protocol, which involves sending embeddings to the active party and receiving backpropagated gradients. The attacker may attempt to infer the labels of the training samples using legitimately obtained data and record the embeddings corresponding to these labels. During the inference phase, the attacker can choose particular source samples for sending malicious embeddings to the active party to facilitate a backdoor attack.

\subsection{Label Inference}

Differing from HFL scenarios, where each participant possesses both features and labels, in VFL scenarios, the passive party only holds features and lacks access to labels. Therefore, inferring the labels is a prerequisite for executing targeted backdoor attacks. To address this, we design the following label inference module.

\begin{figure}[t!]
    \centering
    \subfloat[MNIST]{\includegraphics[width=0.43\columnwidth]{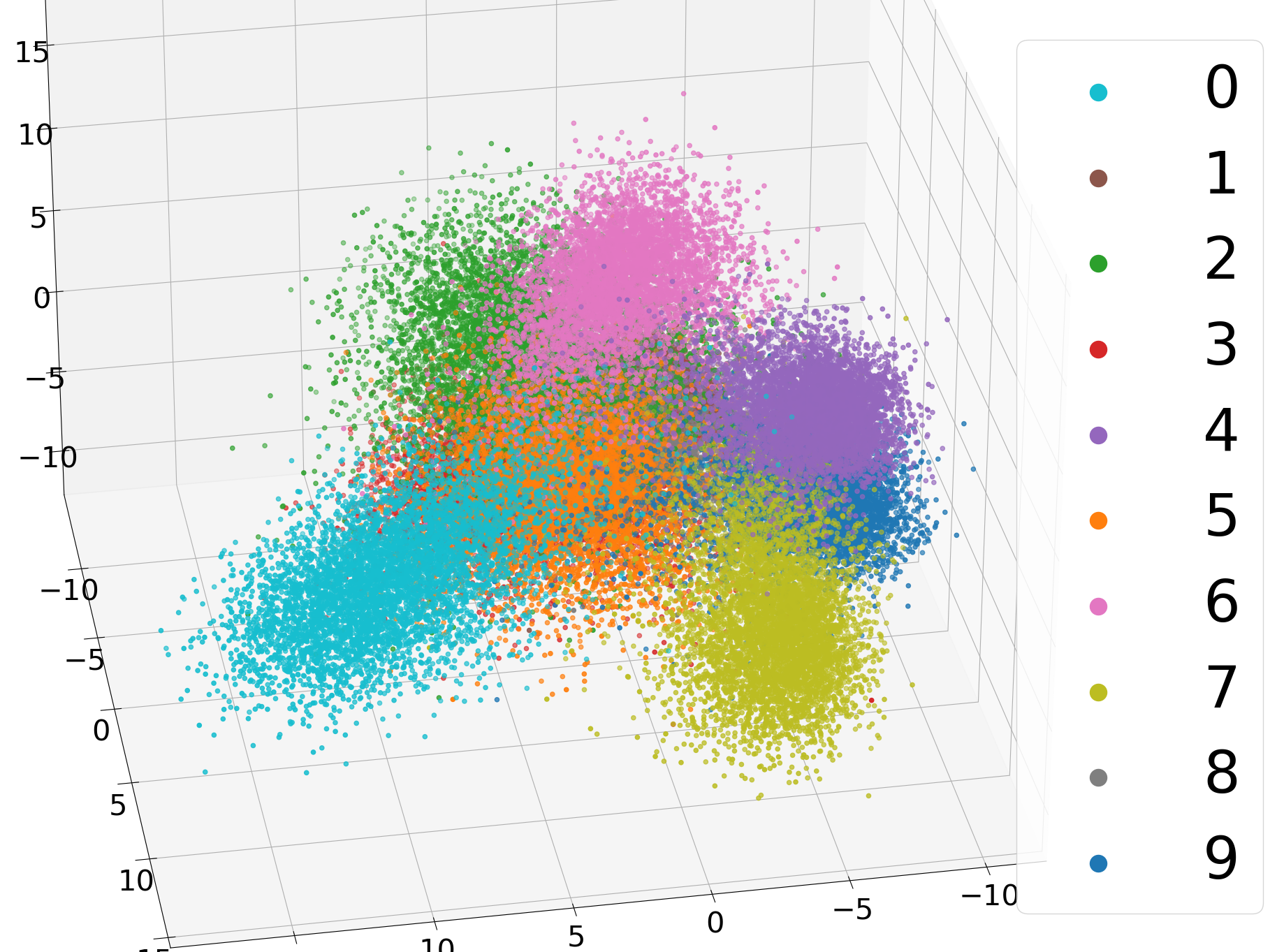}}
    \hspace{2mm}
    \subfloat[CIFAR-10]{\includegraphics[width=0.53\columnwidth]{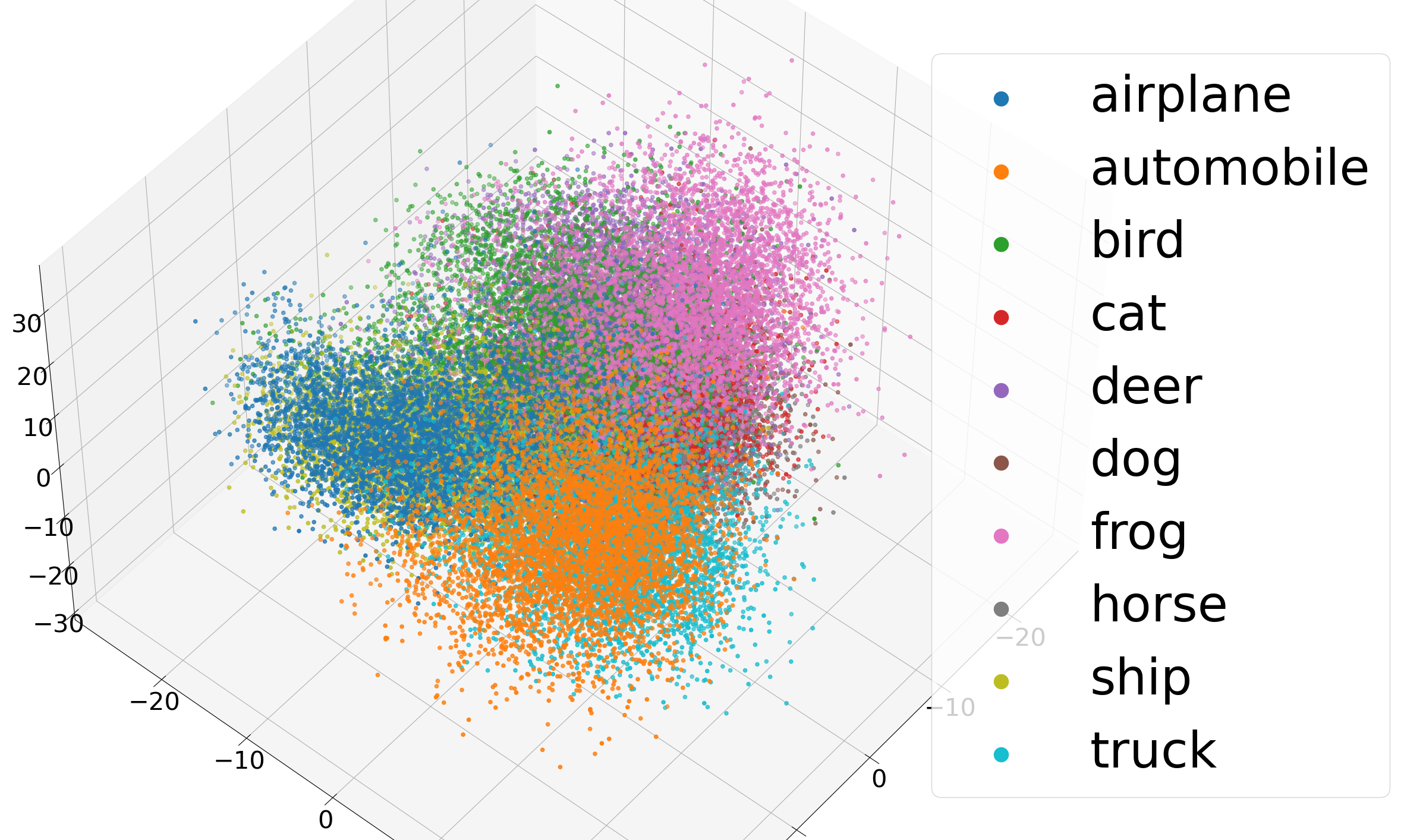}} \\
    \subfloat[FashionMNIST]{\includegraphics[width=0.43\columnwidth]{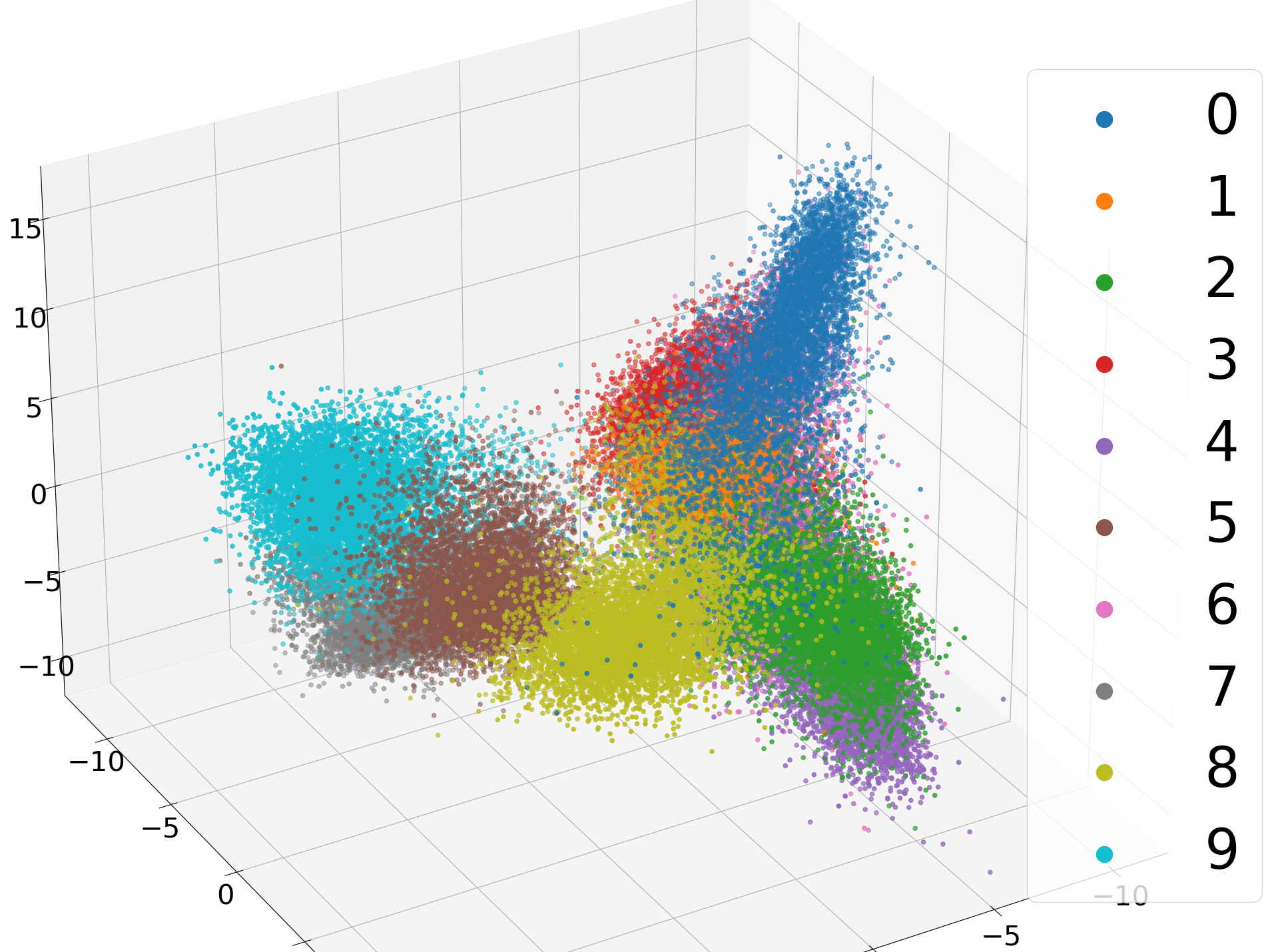}}
    \hspace{2mm}
    \subfloat[CINIC-10]{\includegraphics[width=0.53\columnwidth]{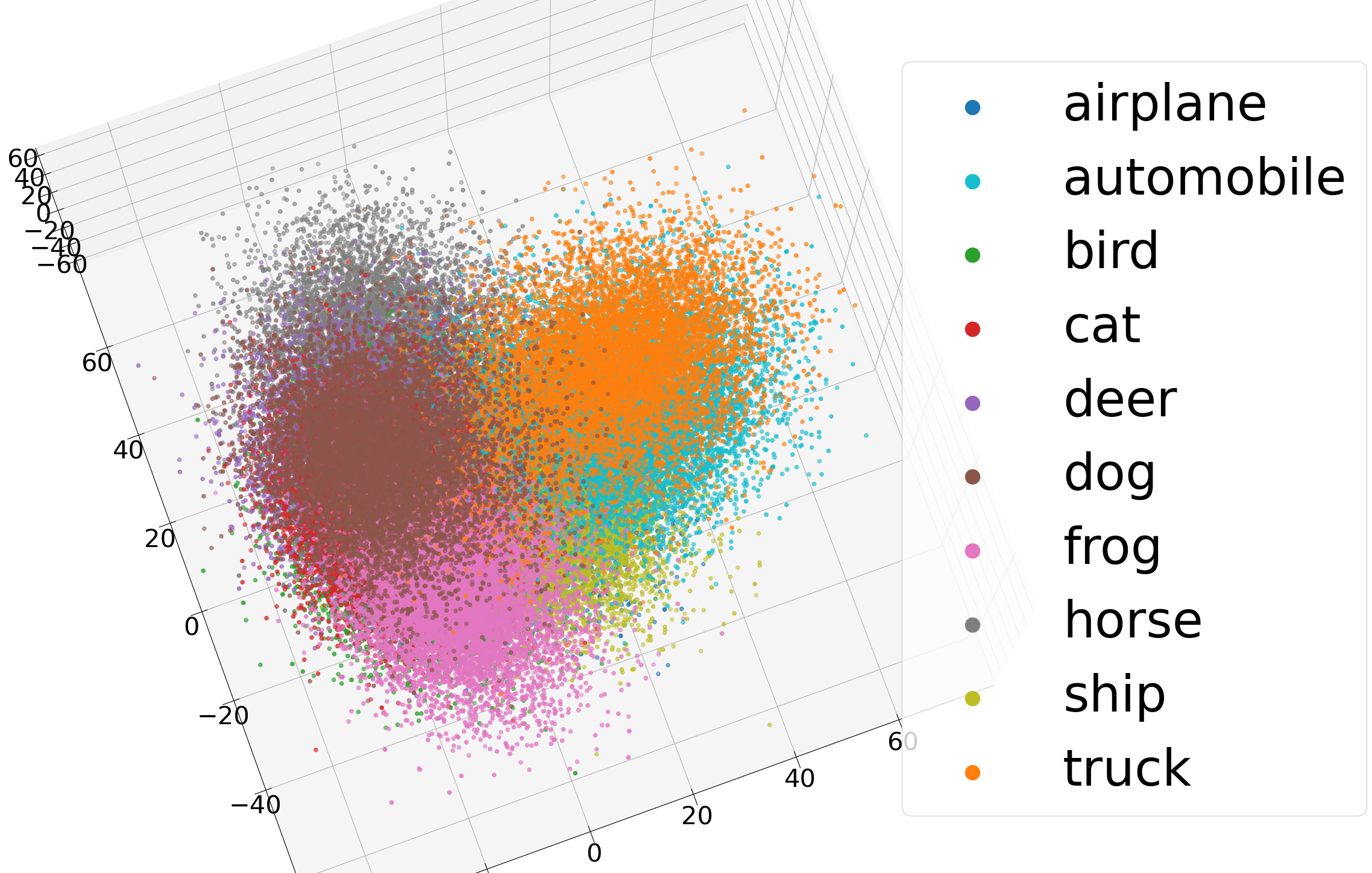}}
    \caption{Distribution of embeddings across the MNIST, FashionMNIST, CIFAR-10, and CINIC-10 datasets, generated as outputs from the passive party's bottom model.}
    \label{fig:emb_distribution}
\end{figure}

We observe that the active party trains the top model using embeddings provided by the passive parties, and these embeddings from a well-trained VFL model effectively capture the corresponding labels. As shown in Fig.~\ref{fig:emb_distribution}, embeddings associated with distinct labels naturally form clusters. Therefore, we employ a clustering algorithm to infer the labels using auxiliary data. Specifically, the attacker begins by selecting a set of embeddings $\left\{\mathrm{emb}_i^a\right\}_{i=1}^N$ recorded during a specific epoch of the training phase. Typically, the attacker prefers to use embeddings from the final epoch, as embeddings closer to the end of training are more effective at characterizing labels. Next, the attacker uses the bottom model $f_a\left(\cdot\right)$ to compute embeddings $Emb^u=f_a\left(X^u\right)$ for the auxiliary data features $X^u$ and calculates the embedding center $\overline{\mathrm{aux}}_i$ for each label category $l_i$ based on the auxiliary data labels $Y^u$. The set of embedding centers derived from auxiliary data for different labels is denoted as $\left\{\overline{\mathrm{aux}}_i\right\}_{i=1}^{|l|}$, where $|l|$ represents the number of label categories. Finally, the attacker applies equation
\begin{equation}
    \left\{\hat{y}_i\right\}_{i=1}^N=\mathrm{Infer}\left(\left\{\mathrm{emb}_i^a\right\}_{i=1}^N,\left\{\overline{\mathrm{aux}}_i\right\}_{i=1}^{|l|},|l|\right)
\end{equation}
to perform clustering analysis and infer the labels. It is worth noting that since both the VFL training task and our label inference approach focus on maximizing the label representation ability of embeddings, the attacker has a high probability of achieving a well-trained model. Therefore, the output embeddings from the bottom model exhibit strong clustering characteristics, rendering the choice of clustering algorithm~(e.g., K-means or mean-shift) relatively inconsequential to label inference accuracy.

\subsection{Poison Generation}

Since the VFL model is distributed among participants, the active party, possessing only a part of the entire model, cannot independently complete the main task during the inference phase. This obviates the need for trigger implantation. Our intuition for implementing backdoor attacks in VFL is to directly replace the original embeddings with those corresponding to the target label. So that the top model of the active party can be misled to make wrong predictions.

To implement this feature-based triggerless backdoor attack, we initially divide the sample embeddings based on the labels inferred in label inference module. Then we calculate the density center $\overline{\mathrm{emb}}_i$ of these embeddings corresponding to each label $l_i$ and denote $\overline{\mathrm{Emb}}=\left\{\overline{\mathrm{emb}}_i\right\}_{i=1}^{|l|}$ as the set of embedding centers, where $\overline{\mathrm{emb}}_t$ represents the embedding center for the target label $l_t$. However, employing only $\overline{\mathrm{emb}}_t$ for backdoor attacks is not sufficient due to the following two significant challenges.

First, in multi-party VFL scenarios with only one attacker, the number of benign participants over attackers renders a single malicious embedding insufficient to mislead the top model. This presents a challenge in amplifying the attacker's malicious embedding to ensure the success of the backdoor attack. To overcome this, we introduce an amplification factor $\eta$ to enhance the malicious embedding
\begin{equation}
    \widehat{\mathrm{emb}}=\eta\cdot\overline{\mathrm{emb}}_t.
\end{equation}
This factor $\eta$ adjusts the contribution of $\overline{\mathrm{emb}}_t$ based on the percentage of attackers among total participants, thereby augmenting the influence of the malicious embedding on the top model to the backdoor attack.

Second, the active party's defense strategy for dynamic detection of sent embeddings is also a challenge. A fixed embedding could be easily identified and eliminated by the active party. To counter this, enhancing the diversity of malicious target embeddings becomes crucial. We address this by adding a perturbation vector that satisfies the Gaussian distribution to $\widehat{\mathrm{emb}}$ and generating a set of malicious embeddings $\widetilde{\mathrm{Emb}}$ that satisfy
\begin{equation}
    \widetilde{\mathrm{emb}}_i=\widehat{\mathrm{emb}}+\delta,\ \delta\sim\mathcal{N}\left(\mu,\sigma^2\right)
\end{equation}
for each $\widetilde{\mathrm{emb}}_i\in\widetilde{\mathrm{Emb}}$. The suitable Gaussian distribution helps ensure that malicious embeddings remain within acceptable prediction boundaries, thus introducing diversity to the embeddings while minimizing the likelihood of failure due to extreme deviations. Therefore, this variation allows us to use the perturbed malicious embeddings to successfully implement the backdoor attack.

\subsection{Backdoor Execution}

Our backdoor attack is implemented during the inference phase, bypassing the need for trigger implantation in the training phase. This triggerless method renders detection-based defenses ineffective, as it does not alter raw features or embeddings during the training phase, and fully complies with the VFL protocol. Furthermore, since the attacker and other participants have the same training objective~(i.e., they all want a well-trained model), a superior model typically results in a more effective backdoor attack.

We design two execution approaches for our triggerless backdoor attack: dirty-label attack and clean-label attack. The dirty-label approach is relatively straightforward: attackers directly replace benign embeddings with malicious embeddings $\widetilde{\mathrm{Emb}}$ generated in the poison generation module to mislead the top model. This approach is similar to traditional trigger-based backdoor attacks, but we replace triggers with malicious embeddings to achieve attacks on any sample. However, in more practical VFL scenarios, where attackers typically aim to induce prediction errors for samples of a specific label (source label $l_s$) toward a target label---achieving a more precise rather than broad attack objective---we propose a more sophisticated clean-label approach. To execute the attack, the attacker initially calculates the embeddings of the samples and infers their labels based on the similarity to the recorded density centers of embeddings $\overline{\mathrm{Emb}}$ corresponding to the different labels. For each sample, based on its corresponding embedding $\mathrm{emb}_i^a$, calculate 
\begin{equation}
    \hat{y}^a_i=\mathrm{argmax}\left(\mathrm{sim}\left(\overline{\mathrm{Emb}},\mathrm{emb}_i^a\right)\right)
\end{equation}
to infer its label. This process helps the attacker to identify whether the samples qualify as source samples. For samples where $\hat{y}^a_i$ equals $l_s$, the attacker replaces their original embeddings with $\widetilde{\mathrm{emb}}_i$ from the set of malicious embeddings $\widetilde{\mathrm{Emb}}$ generated in the second module and sends them to the active party, thereby executing the backdoor attack.

\section{Evaluation}

In this section, we evaluate the attack performance and robustness of our proposed backdoor attack. This includes comparing its attack success rate with three existing baseline attacks, evaluating potential influencing factors, investigating the impact of hyperparameters, and assessing its robustness under various popular defense strategies.

\subsection{Experimental Setup}

We elaborate on our experimental setup, including datasets, models, baselines, and evaluation metrics.

\subsubsection{Datasets} The experiments are conducted on five benchmark datasets: MNIST~(MN)~\cite{lecun1998gradient}, Fashion-MNIST~(FM)~\cite{xiao2017fashion}, CIFAR-10~(CF)~\cite{krizhevsky2009learning}, CINIC-10~(CN)~\cite{darlow2018cinic10}, and Criteo~(CT)~\cite{criteo2014dataset}, where the first four are image datasets and the last is a tabular dataset. We use random sampling along columns to divide features, simulating how passive parties in VFL share the sample space while retaining distinct feature spaces.

\subsubsection{Models} We follow the VFL framework illustrated in Fig.~\ref{fig:backdoor} and conduct experiments in the multi-party VFL scenarios. For the passive party's bottom model, distinct embedding models are set for various datasets. Specifically, a single-layer fully connected network is employed for both MNIST and FashionMNIST datasets. In contrast, a four-layer convolutional neural network is utilized for the CIFAR-10 and CINIC-10 datasets. For the Criteo dataset, the DeepFM model~\cite{guo2017deepfm} is adopted. Regarding the active party, its top model initially aggregates the embeddings forwarded by the passive parties according to their dimensional attributes, and subsequently performs label prediction through a fully connected layer network.

\subsubsection{Baselines} To explore whether triggers are essential in VFL, we compare our backdoor attack with three established methods, each utilizing distinct triggers: VILLAIN~\cite{bai2023villain}, BadVFL~\cite{naseri2023badvfl}, and BASL~\cite{he2023backdoor}. Among these methods, VILLAIN employs strip-based triggers with positive and negative crosses. BadVFL identifies the most contributing features and implants cross triggers in them. BASL uses embedding-based triggers to misdirect embedding predictions. We implement these baselines in the multi-party VFL scenario and evaluate their performance.

\subsubsection{Evaluation Metrics}

We introduce the backdoor implantation rate~(BIR) to evaluate whether the triggers of traditional backdoor attacks~(baselines) are successfully implanted into the VFL model. Since trigger implantation is limited to the training phase, the BIR metric is exclusively applicable during this phase. As for the inference phase, we use the attack success rate~(ASR) to assess the performance of backdoor attacks on the test dataset. Moreover, considering that backdoor implantation and attacks involve label inference, evaluated by the label inference success rate~(LISR), the samples the attacker attempts to manipulate and the samples actually manipulated may differ. Therefore, we refine the above two metrics as follows:
\begin{itemize}
    \item Target BIR~(tBIR): The proportion of samples with target labels correctly predicted as such.
    \item Manipulated BIR~(mBIR): The proportion of samples manipulated by the attacker that are correctly predicted as target labels.
    \item Real BIR~(rBIR): The proportion of manipulated samples, whose real label is the target label, correctly predicted as the target label.
    \item Manipulated ASR~(mASR): The success rate of the backdoor attack on samples manipulated by the attacker.
    \item Real ASR~(rASR): The success rate of the backdoor attack on manipulated samples whose true label is the source label.
\end{itemize}
Given that we have designed two attack execution approaches, we introduce distinct metrics to evaluate their performance. Specifically, dirty-label attacks can be initiated on samples with any label, so the proportion of successfully misclassified samples among all samples manipulated by the attacker represents the backdoor attack performance. To measure this, we introduce mASR as the evaluation metric for dirty-label attacks, aligning with the ``attack success rate'' metric commonly used in existing related works~\cite{naseri2023badvfl,he2023backdoor,bai2023villain}. In contrast, clean-label attacks require simultaneous evaluation of the identification accuracy of source samples and the backdoor attack performance on these samples. Therefore, we introduce the LISR and rASR metrics for a comprehensive evaluation. In addition, we use the main task accuracy~(MTA) to evaluate the impact of the backdoor attack on model training. For all the evaluation metrics described above, higher values represent better performance.

\subsection{Overall Performance}

\begin{table*}[t!]
    \centering
    \caption{Attack performance of our feature-based triggerless method with baselines in the multi-party VFL scenario}
    \label{tab:comparison_passive4}
    \setlength{\tabcolsep}{3.6pt}
    \begin{threeparttable}
        \begin{tabular}{ccccccccc}
            \toprule
            \multirow{2}{*}[-0.4em]{Dataset} & \multirow{2}{*}[-0.4em]{Method} & \multicolumn{3}{c}{BIR} & \multirow{2}{*}[-0.4em]{LISR} & \multicolumn{2}{c}{ASR} & \multirow{2}{*}[-0.4em]{MTA} \\
            \cmidrule{3-5}\cmidrule{7-8}
            & & tBIR & mBIR & rBIR & & mASR & rASR &  \\
            \midrule
            \multirow{4}{*}{MNIST} & VILLAIN & $98.02\pm0.03\%$ & $75.21\pm4.89\%$ & $99.50\pm0.06\%$ & $54.71\pm8.11\%$ & $1.89\pm1.34\%$ & $0.00\pm0.00\%$ & $96.37\pm0.04\%$ \\
            & BadVFL & $98.23\pm0.05\%$ & $75.47\pm5.82\%$ & $99.46\pm0.12\%$ & $54.63\pm7.81\%$ & $2.05\pm0.85\%$ & $0.00\pm0.00\%$ & $96.44\pm0.00\%$ \\
            & BASL & $98.16\pm0.07\%$ & $67.51\pm23.14\%$ & $99.51\pm0.09\%$ & $54.49\pm7.72\%$ & $1.97\pm0.93\%$ & $0.07\pm0.10\%$ & $96.43\pm0.03\%$ \\
            & Our & / & / & / & $55.17\pm7.37\%$ & $\pmb{93.97\pm3.97\%}$ & $\pmb{94.47\pm7.17\%}$ & $\pmb{96.48\pm0.06\%}$ \\
            \midrule
            \multirow{4}{*}{\shortstack{Fashion\\MNIST}} & VILLAIN & $87.01\pm0.20\%$ & $81.27\pm2.82\%$ & $95.97\pm0.84\%$ & $51.46\pm4.79\%$ & $11.57\pm8.34\%$ & $0.12\pm0.09\%$ & $86.43\pm0.04\%$ \\
            & BadVFL & $86.41\pm0.40\%$ & $57.11\pm35.43\%$ & $79.34\pm23.96\%$ & $51.66\pm4.95\%$ & $7.00\pm1.75\%$ & $0.32\pm0.17\%$ & $86.25\pm0.15\%$ \\
            & BASL & $86.27\pm0.12\%$ & $64.97\pm3.18\%$ & $93.03\pm0.56\%$ & $51.29\pm5.12\%$ & $7.11\pm0.60\%$ & $0.21\pm0.17\%$ & $86.45\pm0.08\%$ \\
            & Our & / & / & / & $51.39\pm4.92\%$ & $\pmb{86.06\pm8.83\%}$ & $\pmb{81.22\pm11.64\%}$ & $\pmb{86.48\pm0.08\%}$ \\
            \midrule
            \multirow{4}{*}{CIFAR-10} & VILLAIN & $78.64\pm1.16\%$ & $20.49\pm14.62\%$ & $72.06\pm14.81\%$ & $21.34\pm0.64\%$ & $7.85\pm1.40\%$ & $2.73\pm1.94\%$ & $65.04\pm1.33\%$ \\
            & BadVFL & $73.89\pm1.09\%$ & $39.31\pm2.70\%$ & $85.23\pm3.29\%$ & $20.46\pm2.59\%$ & $9.19\pm11.23\%$ & $0.00\pm0.00\%$ & $64.76\pm1.57\%$ \\
            & BASL & $78.01\pm1.45\%$ & $13.69\pm13.13\%$ & $94.28\pm8.09\%$ & $21.38\pm1.52\%$ & $9.14\pm5.60\%$ & $0.00\pm0.00\%$ & $64.75\pm1.54\%$ \\
            & Our & / & / & / & $20.72\pm0.97\%$ & $\pmb{85.08\pm11.38\%}$ & $\pmb{83.99\pm12.81\%}$ & $\pmb{65.15\pm1.16\%}$ \\
            \midrule
            \multirow{4}{*}{CINIC-10} & VILLAIN & $77.23\pm0.30\%$ & $37.80\pm8.39\%$ & $85.99\pm2.81\%$ & $20.65\pm2.55\%$ & $0.00\pm0.00\%$ & $0.00\pm0.00\%$ & $56.12\pm0.90\%$ \\
            & BadVFL & $76.27\pm0.42\%$ & $45.93\pm7.26\%$ & $89.06\pm1.48\%$ & $19.77\pm3.60\%$ & $1.20\pm1.70\%$ & $0.00\pm0.00\%$ & $53.93\pm0.95\%$ \\
            & BASL & $78.52\pm0.45\%$ & $20.21\pm20.83\%$ & $57.61\pm41.05\%$ & $20.94\pm1.22\%$ & $4.88\pm4.12\%$ & $2.74\pm2.77\%$ & $56.47\pm0.65\%$ \\
            & Our & / & / & / & $20.98\pm0.95\%$ & $\pmb{88.54\pm13.22\%}$ & $\pmb{92.72\pm6.86\%}$ & $\pmb{56.53\pm0.17\%}$ \\
            \midrule
            \multirow{4}{*}{Criteo} & VILLAIN & $99.81\pm0.05\%$ & $99.77\pm0.08\%$ & $99.83\pm0.05\%$ & $64.41\pm2.94\%$ & $99.96\pm0.05\%$ & $99.95\pm0.08\%$ & $77.54\pm0.01\%$ \\
            & BadVFL & $99.72\pm0.26\%$ & $99.64\pm0.32\%$ & $99.73\pm0.24\%$ & $65.09\pm2.22\%$ & $99.92\pm0.11\%$ & $99.90\pm0.13\%$ & $77.51\pm0.03\%$ \\
            & BASL & $99.74\pm0.02\%$ & $99.67\pm0.03\%$ & $99.76\pm0.03\%$ & $64.92\pm1.84\%$ & $99.90\pm0.04\%$ & $99.86\pm0.04\%$ & $77.53\pm0.00\%$ \\
            & Our & / & / & / & $65.48\pm2.66\%$ & $\pmb{100.00\pm0.00\%}$ & $\pmb{100.00\pm0.00\%}$ & $\pmb{77.55\pm0.00\%}$ \\
            \bottomrule
        \end{tabular}
        \begin{tablenotes}
            \footnotesize
            \item[] The \textbf{bold} data indicate the best performance.
        \end{tablenotes}
    \end{threeparttable}
\end{table*}

We evaluate our attack in a more challenging multi-party VFL scenario rather than the two-party scenario~(results for the two-party scenario are provided in Appendix~B). This scenario contains one active party and multiple passive parties: three passive parties for the Criteo dataset and four passive parties for the other datasets. We compare it with three existing backdoor attacks across five datasets. In the experiments, the batch size is set to 128 for all datasets except CIFAR-10, which is set to 64. Importantly, all experiments use only one set of auxiliary dataset, meaning the attacker knows only one sample in each class. As shown in Table~\ref{tab:comparison_passive4}, our proposed feature-based triggerless backdoor attack outperforms in attack success rate while minimally impacting the main task. This result indicates that triggers, traditionally considered crucial in backdoor attacks, may not be indispensable within VFL contexts.

Moreover, the evaluation results on LISR show that our attack maintains outstanding backdoor attack performance even with general label inference capabilities. This is primarily because, unlike traditional attacks that rely on the backdoor implantation process, our attack performance is driven by the inherent capability of embeddings to represent labels rather than by label inference accuracy. Specifically, in VFL scenarios where attackers lack label knowledge, traditional attacks require label inference to identify target samples and implant triggers within them to execute targeted backdoor attacks. However, due to the limitations of label inference accuracy, attackers inevitably implant triggers into non-target samples, which degrades the overall backdoor attack performance. In contrast, our triggerless attack directly replaces original embeddings with malicious embeddings during the inference phase. The generation of these malicious embeddings relies primarily on the natural correlation between embeddings and labels. Guided by labeled auxiliary data, malicious embeddings are created by calculating cluster centers based on the natural clustering of embeddings. As a result, our attack is minimally influenced by the performance of label inference.

In addition, while conventional baseline attacks perform well in two-party VFL scenarios, they perform significantly worse than our attack in multi-party VFL scenarios. Typically, the ASR disparity between our attack and other backdoor attacks can reach up to 50 times. We attribute this significant disparity to the following reasons. First, due to the limited label inference ability, triggers are poorly implanted in the target labeled samples. For instance, in the MNIST dataset, the tBIR exceeding 98\% signifies accurate prediction for most target samples, and the rBIR above 99\% confirms successful backdoor implantation. However, the mBIR being around 75\% reveals that many backdoor-implanted samples do not match target samples. Second, conventional backdoor attacks implant triggers only in the training set. Distributional differences between training and test sets cause poor performance on the test set. Third, conventional attacks need to evade the active party's defense detection during training, resulting in limited changes to the embeddings with triggers. The distributional disparity between test and training sets may cause larger embedding differences than those caused by triggers, leading to attack failure. Fourth, in multi-party scenarios, benign passive parties dilute the attacker's malicious embeddings, reducing the impact of backdoor triggers on the overall VFL model and amplifying attack failure.

We also observe that our proposed triggerless backdoor attack achieves the best MTA across all five datasets compared to traditional backdoor attacks, with minimal impact on the model's decision boundaries. This is primarily because our backdoor attack does not perform any operations that violate the training protocol during the training phase, except for recording the training data. In other words, the decision boundaries of the models obtained through our attack during the training phase remain identical to those from benign training, unaffected by poisoned data, thereby preserving the model's predictive performance. In contrast, traditional backdoor attacks that involve trigger implantation use poisoned data to mislead the training process, causing the model to learn incorrect distributional information. This alters the model's decision boundaries and compromises its predictive performance. Additionally, since the accuracy of our attack heavily depends on the performance of the bottom model, the attacker's training goal aligns with that of other benign participants---namely, achieving a well-trained bottom model. This shared goal further incentivizes our backdoor attack to maintain the model's strong predictive ability during the training phase. We believe that these findings further highlight the advantages of our proposed triggerless backdoor attack compared to traditional backdoor attacks.

\subsubsection{Impact of the Number of Passive Parties}

\begin{table}[t!]
    \centering
    \caption{Attack performance with different numbers of passive parties}
    \label{tab:num_pp}
    \begin{threeparttable}
        \begin{tabular}{cccc}
            \toprule
            Dataset & \#PP\tnote{$\ast$} & mASR & rASR \\
            \midrule
            \multirow{6}{*}{MNIST} & 1 & $100.00\pm0.00\%$ & $100.00\pm0.00\%$ \\
            & 2 & $97.23\pm2.67\%$ & $98.02\pm1.84\%$ \\
            & 4 & $93.97\pm3.97\%$ & $94.47\pm7.17\%$ \\
            & 7 & $84.11\pm8.12\%$ & $88.45\pm8.94\%$ \\
            & 14 & $77.66\pm17.78\%$ & $80.77\pm16.22\%$ \\
            & 28 & $70.78\pm14.20\%$ & $81.62\pm14.88\%$ \\
            \midrule
            \multirow{6}{*}{CINIC-10} & 1 & $100.00\pm0.00\%$ & $100.00\pm0.00\%$ \\
            & 2 & $94.29\pm1.18\%$ & $92.26\pm1.12\%$ \\
            & 4 & $91.87\pm8.58\%$ & $92.72\pm6.86\%$ \\
            & 8 & $77.92\pm13.08\%$ & $81.42\pm16.22\%$ \\
            & 16 & $75.14\pm10.04\%$ & $73.32\pm17.00\%$ \\
            & 32 & $54.25\pm4.25\%$ & $65.38\pm15.38\%$ \\
            \bottomrule
        \end{tabular}
        \begin{tablenotes}
            \footnotesize
            \item[$\ast$] \#PP denotes the number of passive parties.
        \end{tablenotes}
    \end{threeparttable}
\end{table}

To demonstrate that our triggerless backdoor attack can handle complex multi-party VFL scenarios, we evaluate the impact of the number of passive parties on our attack. As shown in Table~\ref{tab:num_pp}, the ASR of our attack does not decrease proportionally with the increase in the number of passive parties but largely maintains high performance, even though its variance may increase. We believe the main reason for this sustained performance is the introduction of the amplification factor. This factor tunes the embeddings, allowing the aggregated embeddings to maintain the characteristics of the target label while avoiding dilution by the benign embeddings of other passive parties. This result demonstrates the robustness of our attack in multi-party VFL.

\subsubsection{Impact of Auxiliary Data}

\begin{figure}[t]
    \centering
    \includegraphics[width=0.42\columnwidth]{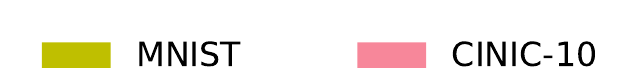}\\
    \subfloat[mASR]{\includegraphics[width=0.48\columnwidth]{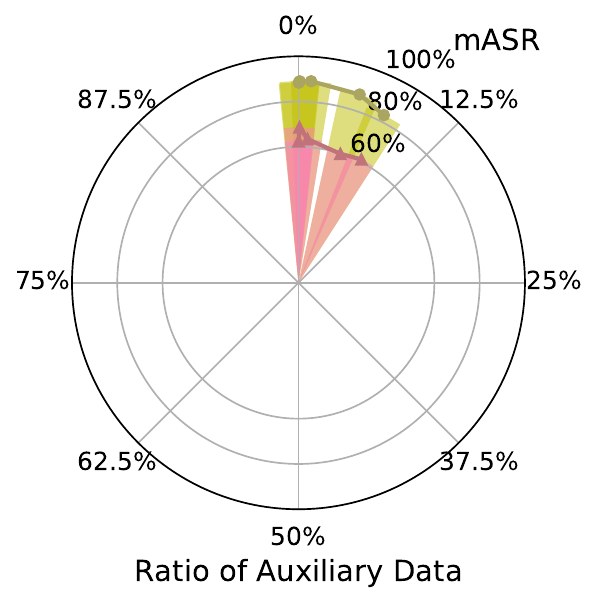}}    
    \hspace{0.02\columnwidth}
    \subfloat[rASR]{\includegraphics[width=0.48\columnwidth]{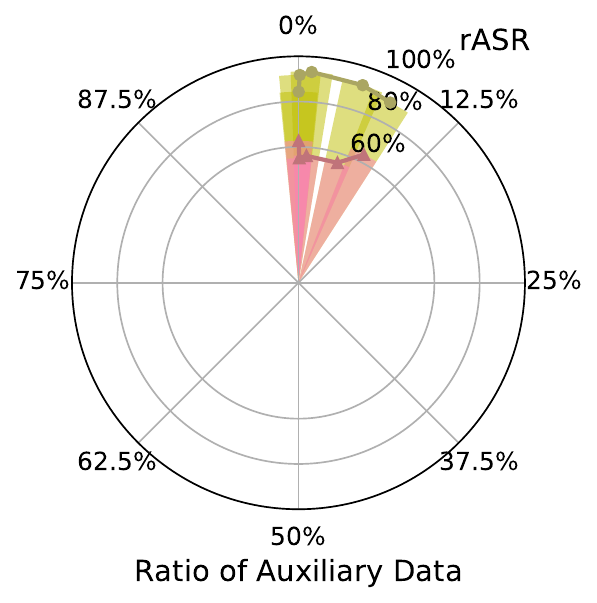}}
    \caption{Impact of the ratio of the auxiliary data on the ASR. The polar radius represents the attack success rate (mASR or rASR), while the polar angle represents the percentage of auxiliary data relative to the training data.}
    \label{fig:auxiliary}
\end{figure}

\begin{figure}[t]
    \centering
    \includegraphics[width=\columnwidth]{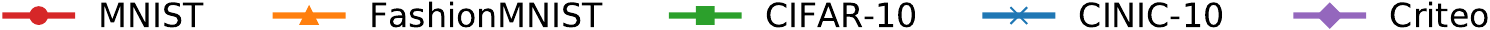}\\
    \subfloat[mASR]{\includegraphics[width=0.48\columnwidth]{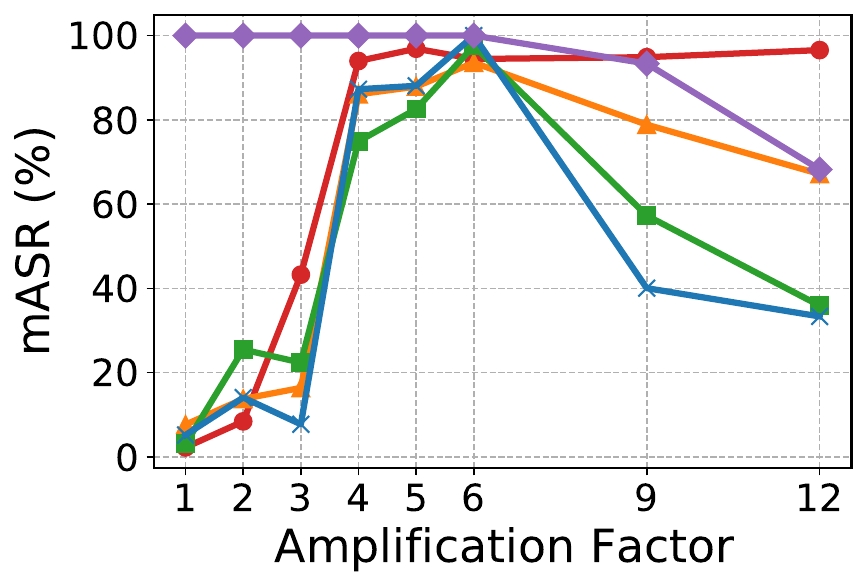}}    
    \hspace{2mm}
    \subfloat[rASR]{\includegraphics[width=0.48\columnwidth]{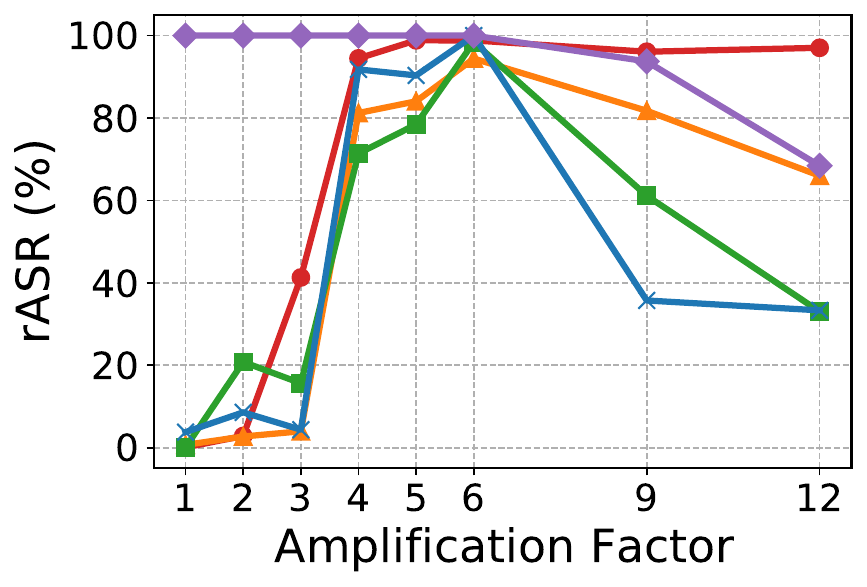}}
    \caption{Impact of the amplification factor on the ASR. In this experiment, there are four passive parties: three benign and one attacker.}
    \label{fig:amplification}
\end{figure}

Our attack uses auxiliary data to identify source and target labels and obtain their clustering features, making it important to evaluate the impact of the amount of auxiliary data on our attack. Therefore, we assess auxiliary data of different sizes on the MNIST and CINIC-10 datasets with four passive parties. We define a set of auxiliary data as one instance of each class. We test the impact on backdoor attacks with 1, 6, 60, 300, and 450 sets of auxiliary data, which account for 0.017\%, 0.167\%, 1.667\%, 5\%, and 7.5\% of the entire dataset, respectively. As shown in Fig.~\ref{fig:auxiliary}, we find that the amount of auxiliary data has a limited impact on the performance of backdoor attacks, although more auxiliary data typically results in better attack performance. Additionally, the experimental results show that even with only one set of auxiliary data, our attack maintains a good attack success rate, which reflects the robustness of our attack. We attribute this to the label characterization ability of the attacker's bottom model. The shared goal between the attacker and other participants~(obtaining a model with good predictive power) enables the attacker to likely obtain a bottom model with strong label characterization ability, thereby facilitating the triggerless backdoor attack.

\begin{figure}[t]
    \centering
    \includegraphics[width=0.63\columnwidth]{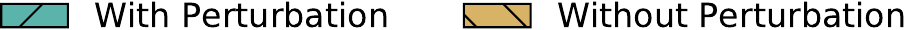}\\
    \subfloat[mASR]{\includegraphics[width=0.48\columnwidth]{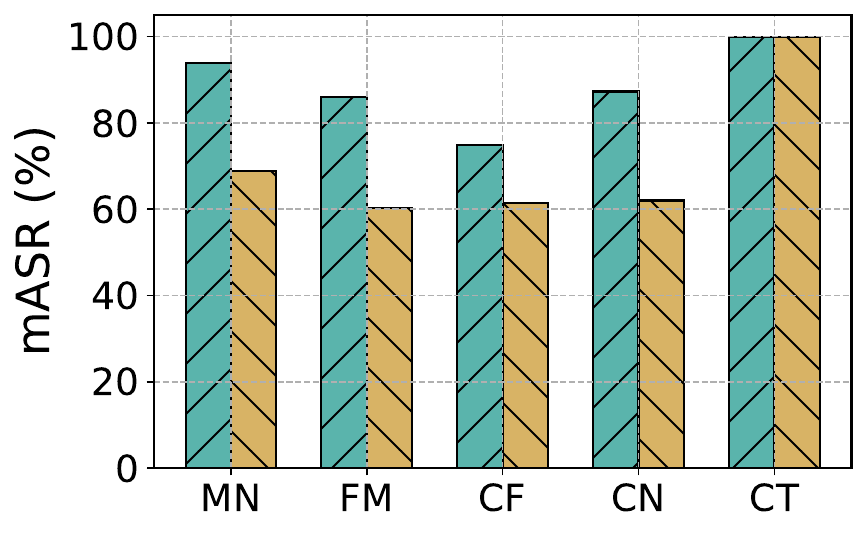}}    
    \hspace{2mm}
    \subfloat[rASR]{\includegraphics[width=0.48\columnwidth]{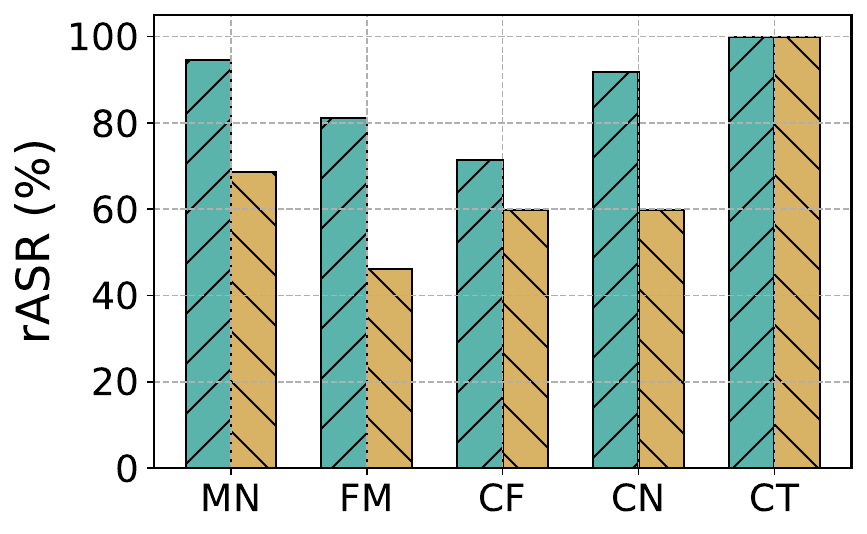}}
    \caption{Impact of the perturbation on the ASR.}
    \label{fig:perturbation}
\end{figure}

\begin{figure}[t]
    \centering
    \subfloat[MNIST]{\includegraphics[width=0.48\columnwidth]{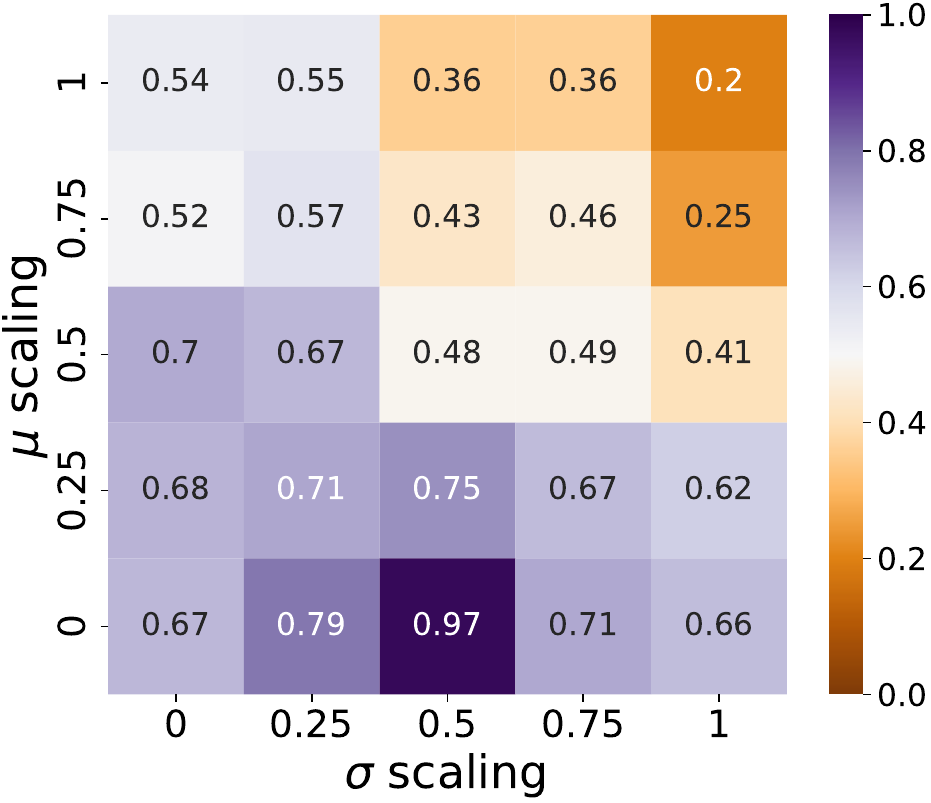}}    
    \hspace{2mm}
    \subfloat[CIFAR-10]{\includegraphics[width=0.48\columnwidth]{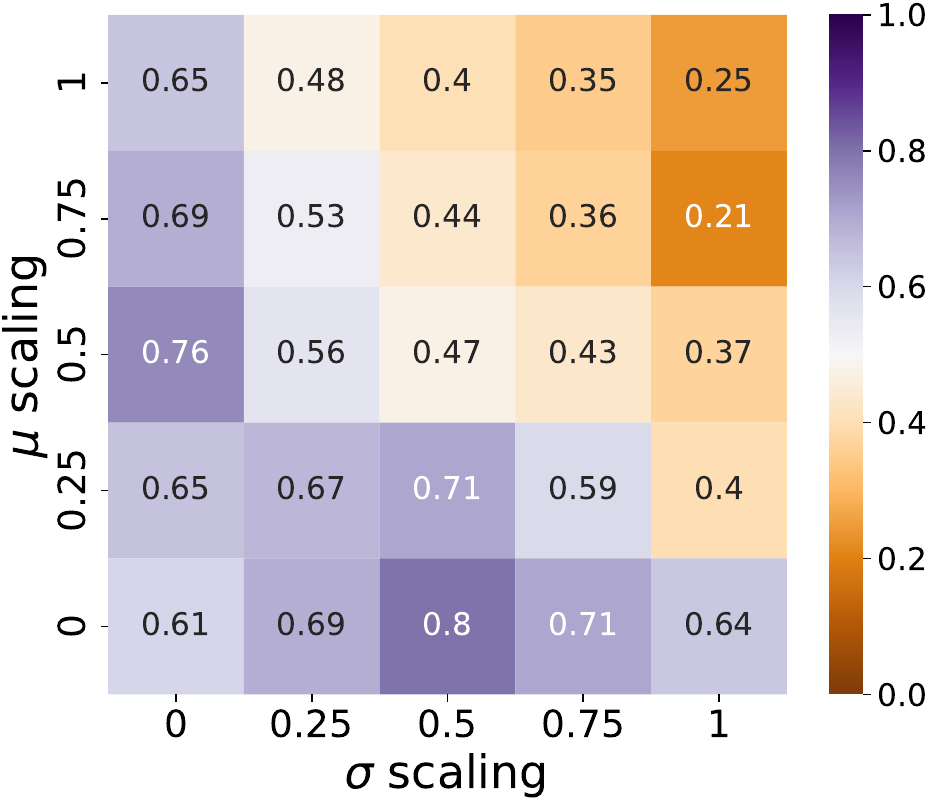}}
    \caption{Impact of the Gaussian distribution on the ASR.}
    \label{fig:gaussian}
\end{figure}

\subsubsection{Impact of Amplification Factor} We investigate the impact of the amplification factor $\eta$ on attack performance. The experiment is conducted in a VFL scenario with three benign passive parties and one attacker. As shown in Fig.~\ref{fig:amplification}, the attack success rate increases with the amplification factor. However, when the amplification factor exceeds a certain threshold, the ASR reaches a high point and slows down its growth rate, suggesting that the effect of the amplification factor on the ASR remains constant beyond this point. We investigate this turning point in attack performance by assuming that $\eta_0$ represents the ratio of benign passive parties to malicious passive parties, and we find that the performance turning point mostly occurs after $\eta_0$. We attribute this phenomenon to the influence of the amplification factor on the contribution of the malicious embedding to the top model. In addition, a larger amplification factor is not always better. We observe that the performance of an attack typically decreases significantly when the magnification factor exceeds $2\eta_0$. This occurs because an excessively large amplification factor increases the likelihood of being detected by defense strategies during the prediction phase and may cause the backdoor attack to produce non-targeted incorrect results. Therefore, we suggest choosing an amplification factor slightly larger than $\eta_0$ for the attack.

\subsubsection{Impact of Perturbation}
We evaluate the impact of the perturbation mechanism on attack performance on five datasets, as shown in Fig.~\ref{fig:perturbation}. Our experimental results on the perturbation mechanism demonstrate that it not only reduces the probability of the attack being detected by the active party but also significantly increases the attack success rate. We believe the primary reason for this increase is that the moderate introduction of perturbation enhances the diversity of malicious embeddings while partially correcting errors caused by non-targeted labeled embeddings. However, excessive perturbation may lead to a decrease in attack success rate. This is because the perturbation mechanism may introduce noise that affects the clustering of embeddings, thereby reducing the attack performance. To determine the optimal magnitude of Gaussian perturbation, we evaluate the impact of perturbations with varying Gaussian distributions on the success rate of our attacks across MNIST and CIFAR-10 datasets. The results are presented in Fig.~\ref{fig:gaussian}. The terms ``$\mu$ scaling'' and ``$\sigma$ scaling'' refer to the scaling of the expectation~($\mu$) and standard deviation~($\sigma$) of the Gaussian noise relative to the embedding mean and standard deviation derived from the auxiliary data, respectively. Fig.~\ref{fig:gaussian} shows the results using mASR as the metric, while the results using rASR are provided in Appendix~B. We observe that when the expectation of the Gaussian distribution for the added perturbation is 0, the perturbation consistently yields a positive impact, achieving the best attack performance at a $\sigma$ scaling of 0.5. This occurs because the added perturbation minimally alters the embedding center, allowing the small amount of noise introduced to compensate for errors caused by the auxiliary data and enhance the attack's success rate. Conversely, as $\mu$ scaling increases, the Gaussian perturbation begins to resemble the effect of increasing the amplification factor, resulting in reduced attack performance. Similarly, higher $\sigma$ scaling introduces more outliers, causing a shift in the malicious embeddings and negatively affecting attack performance. Based on these findings, we recommend adding Gaussian perturbation with a expectation of 0 and a standard deviation equal to half the standard deviation of the corresponding embedding distribution derived from the auxiliary data to enhance the diversity of malicious embeddings and improve the attack success rate.

\subsubsection{Impact of Label Categories}

\begin{figure}[t]
    \centering
    \includegraphics[width=0.3\columnwidth]{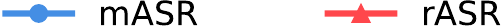}\\
    \subfloat[Source Label]{\includegraphics[width=0.48\columnwidth]{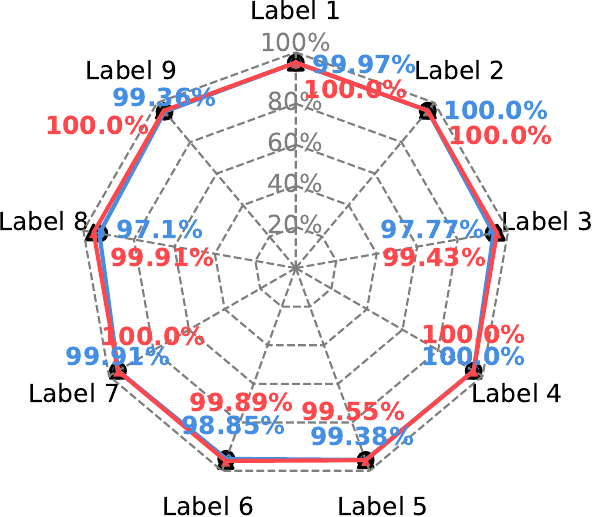}}
    \hspace{0.02\columnwidth}
    \subfloat[Target Label]{\includegraphics[width=0.48\columnwidth]{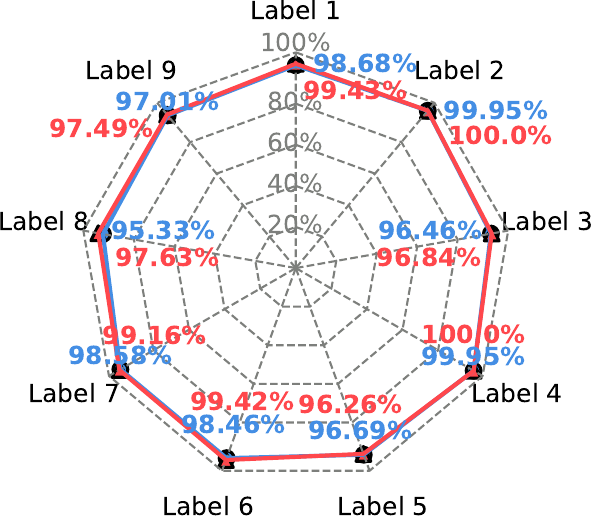}}
    \caption{Attack performance under different source and target label categories.}
    \label{fig:label_categories}
\end{figure}

Fig.~\ref{fig:label_categories} shows the performance of our attack when it is implemented using different source and target label categories, respectively. The evaluation is conducted on the MNIST dataset in a two-party VFL scenario. To analyze the impact of different source labels on the attack success rate, the target label is fixed at 0. Similarly, to examine the effect of different target labels on the attack success rate, the source label is fixed at 0. The results demonstrate that our attack is largely independent of the label categories.

\subsection{Hyperparameter Analysis}

To comprehensively evaluate our triggerless backdoor attack, we investigate and analyze the impact of various hyperparameters on attack performance in detail below. Their combined impact is presented in Appendix~B.

\subsubsection{Impact of Learning Rate}

\begin{figure*}[t]
    \centering
    \includegraphics[width=0.16\textwidth]{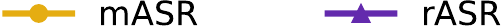}\\
	\subfloat[MNIST]{\includegraphics[width=0.23\textwidth]{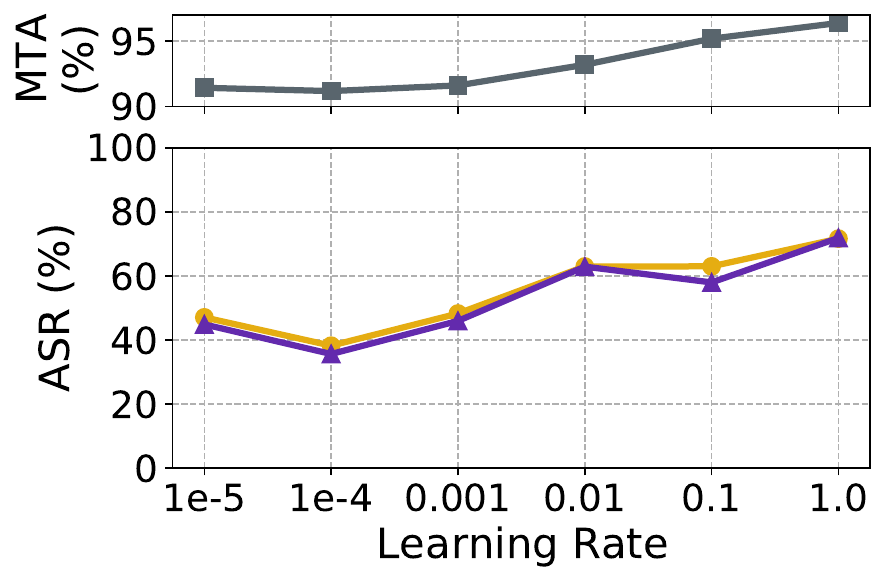}}
    \hspace{2mm}
    \subfloat[FashionMNIST]{\includegraphics[width=0.23\textwidth]{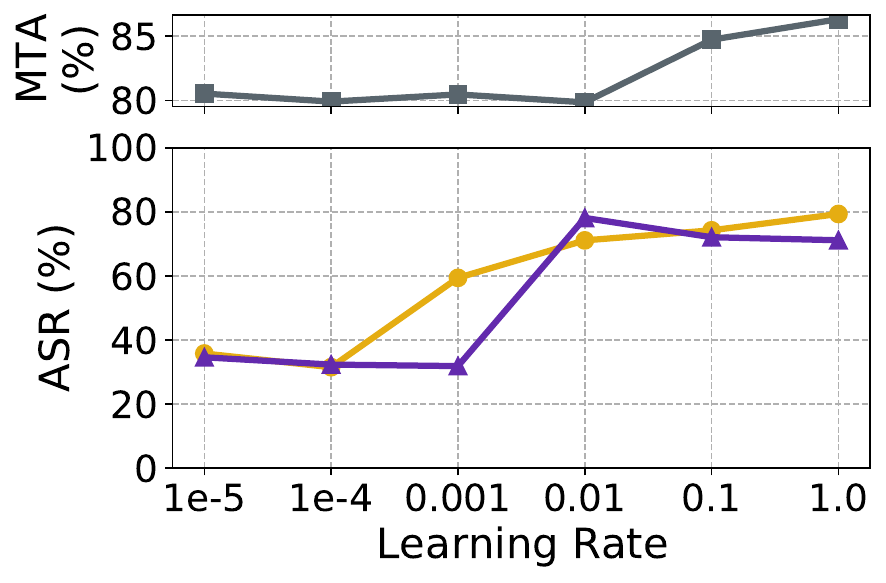}}
    \hspace{2mm}
    \subfloat[CIFAR-10]{\includegraphics[width=0.23\textwidth]{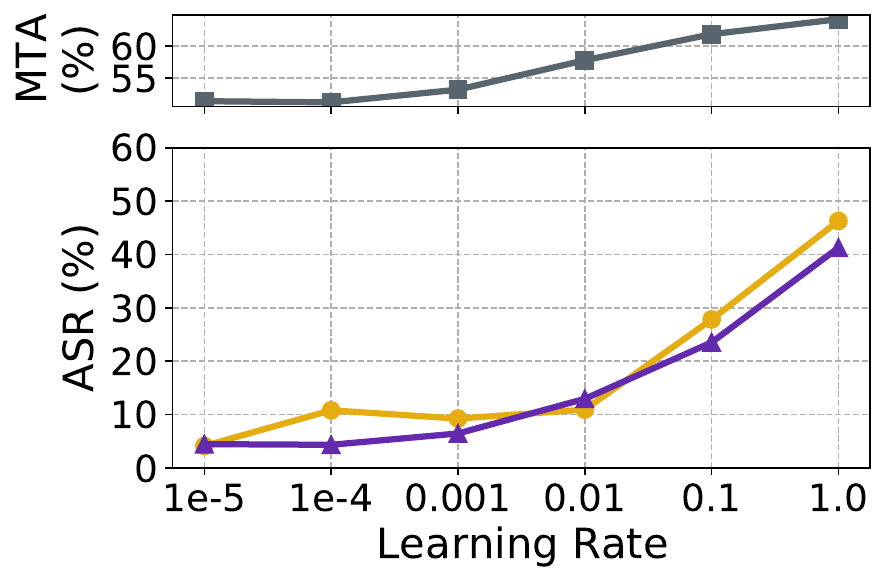}}
    \hspace{2mm}
    \subfloat[CINIC-10]{\includegraphics[width=0.23\textwidth]{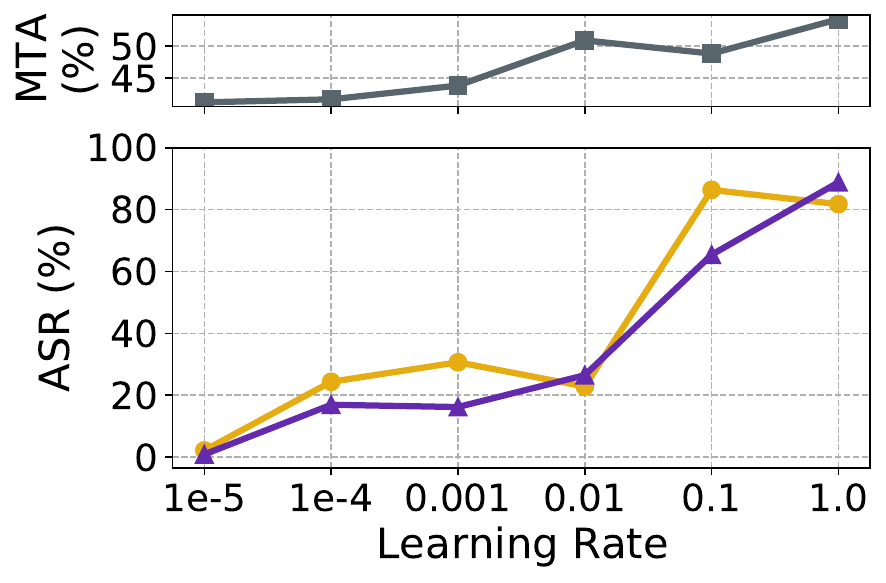}}
    \caption{Impact of learning rate on attack performance.}
    \label{fig:learning_rate}
\end{figure*}

We evaluate the impact of the learning rate on our attack across four datasets. The results are shown in Fig.~\ref{fig:learning_rate}. We observe that for the same number of training epochs, the MTA increases with the learning rate. Since the effectiveness of our backdoor attack relies heavily on the label characterization ability of the bottom model, a higher MTA typically indicates a better model. Consequently, the accuracy of our attack increases with the learning rate. However, a higher learning rate is not always better, as it does not always lead to a better model. Additionally, an excessively high learning rate may introduce additional computational and communication overhead.

\subsubsection{Impact of Batch Size}

\begin{figure*}[t]
    \centering
    \includegraphics[width=0.16\textwidth]{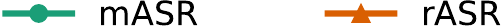}\\
	\subfloat[MNIST]{\includegraphics[width=0.23\textwidth]{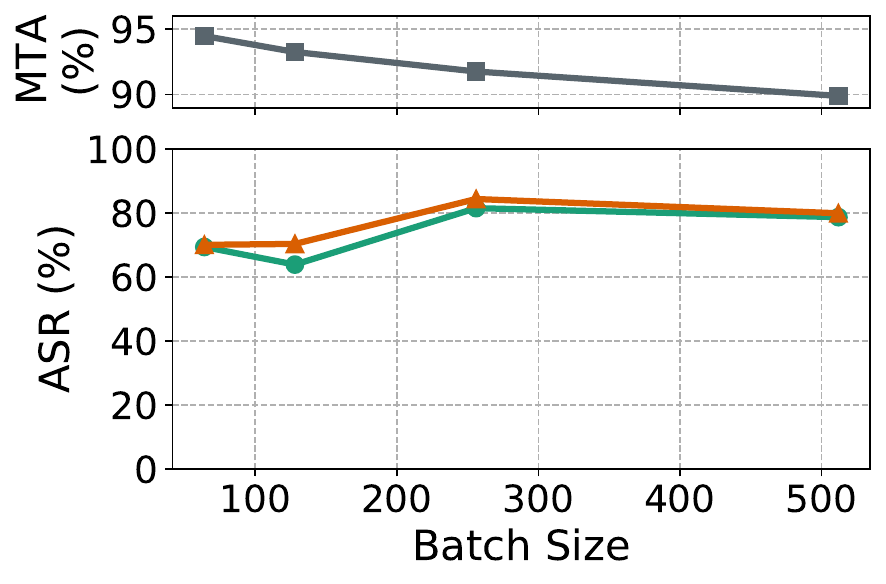}}
    \hspace{2mm}
    \subfloat[FashionMNIST]{\includegraphics[width=0.23\textwidth]{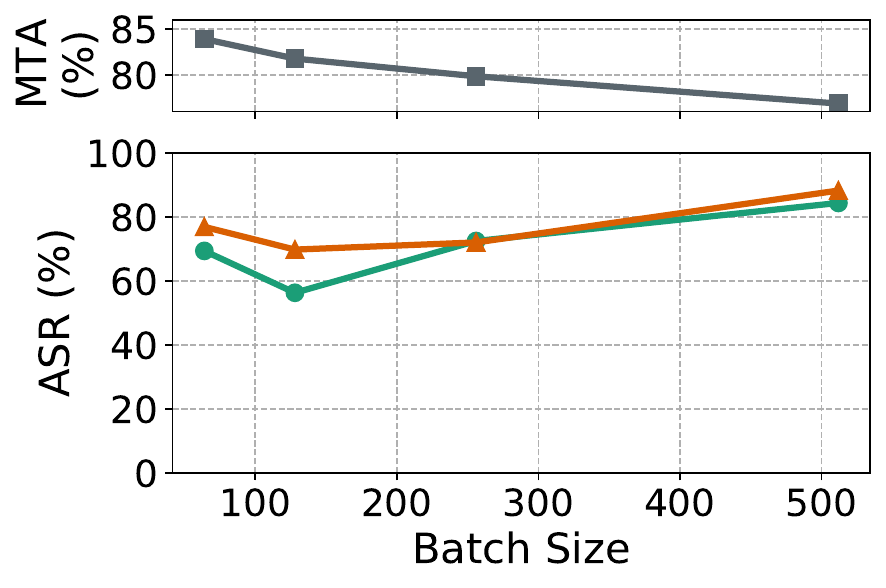}}
    \hspace{2mm}
    \subfloat[CIFAR-10]{\includegraphics[width=0.23\textwidth]{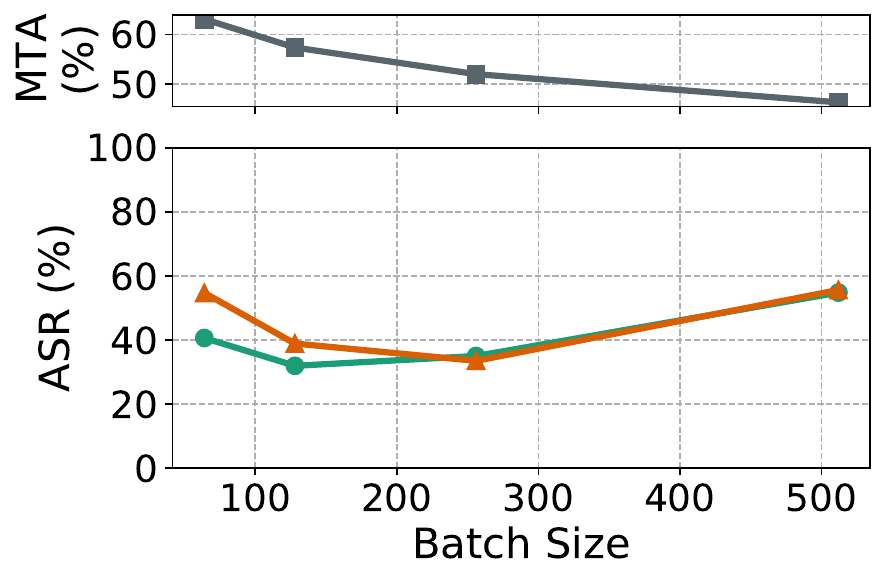}}
    \hspace{2mm}
    \subfloat[CINIC-10]{\includegraphics[width=0.23\textwidth]{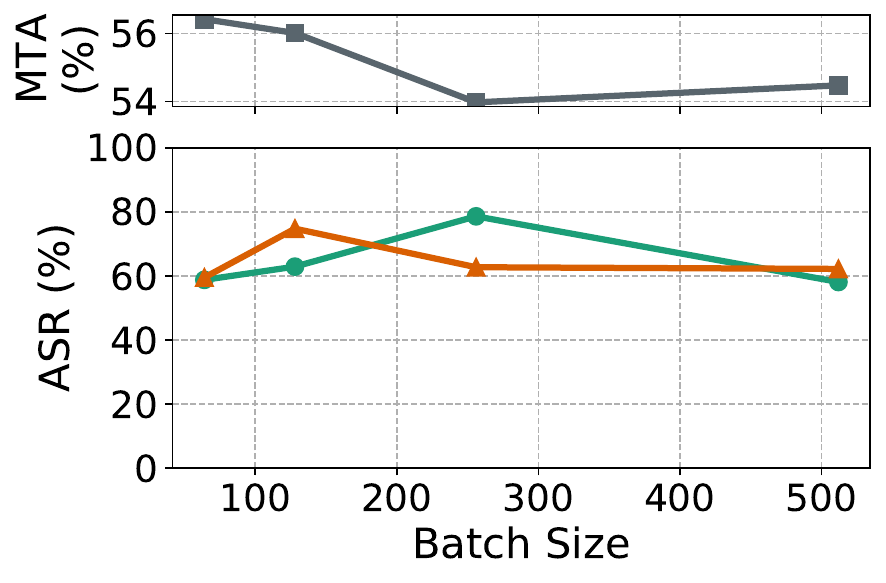}}
    \caption{Impact of batch size on attack performance.}
    \label{fig:batch_size}
\end{figure*}

The batch size setting during the training phase is also an important factor. We evaluate the performance of our backdoor attack on four datasets. As shown in Fig.~\ref{fig:batch_size}, the experimental results indicate that batch size does not have a significant trending effect on attack accuracy, causing the ASR to fluctuate within a certain range. With the same training epochs, the MTA shows a decreasing trend as batch size increases, but the attack performance does not change significantly. Therefore, we conclude that batch size is not a major factor affecting the performance of our triggerless backdoor attack.

\subsubsection{Impact of Training Epochs}

\begin{figure}[t]
    \centering
    \includegraphics[width=0.33\columnwidth]{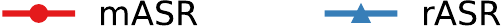}\\
    \subfloat[CIFAR-10]{\includegraphics[width=0.48\columnwidth]{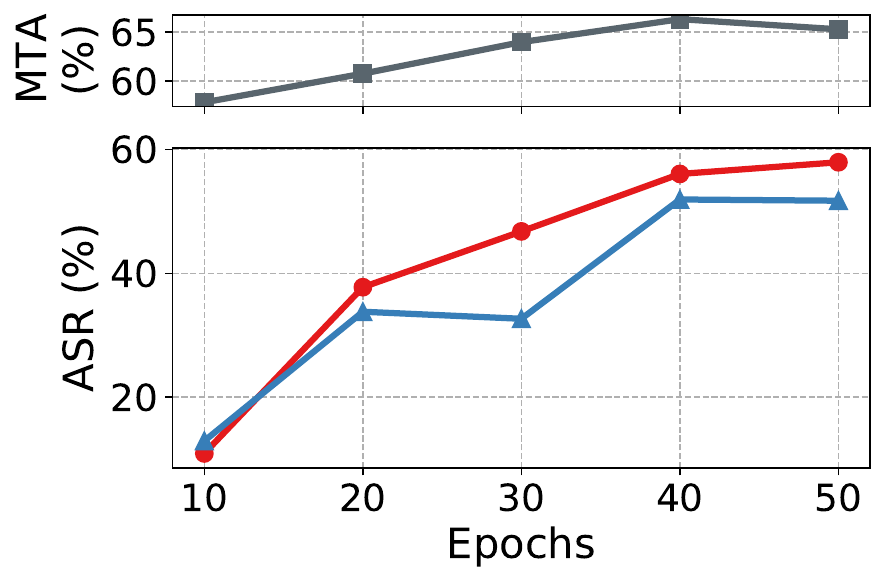}}    
    \hspace{0.02\columnwidth}
    \subfloat[CINIC-10]{\includegraphics[width=0.48\columnwidth]{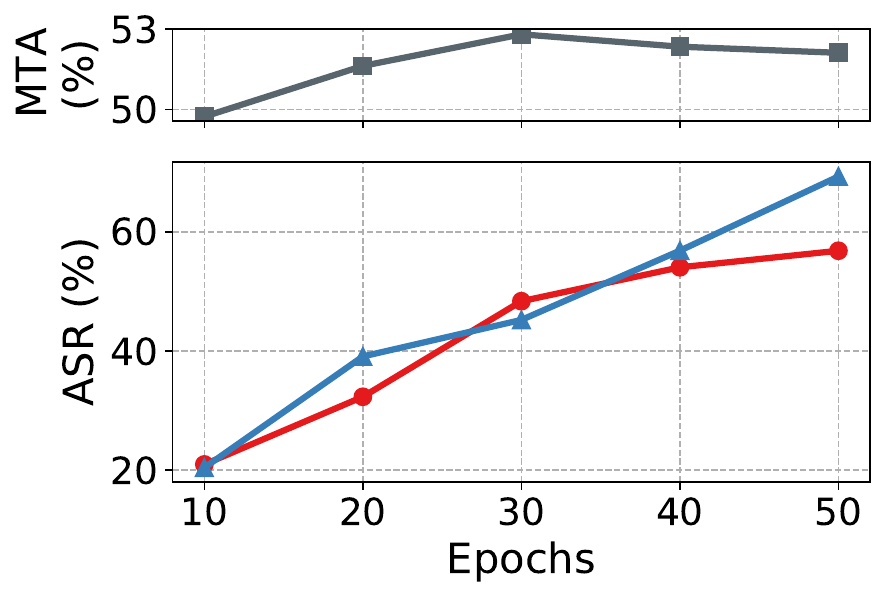}}
    \caption{Impact of training epochs on attack performance.}
    \label{fig:epochs}
\end{figure}

The impact of training epochs on the accuracy of our backdoor attack is evaluated on CIFAR-10 and CINIC-10, as the MTA of these datasets is more sensitive to changes in epochs, better reflecting the impact of different model performances on the triggerless backdoor attack. As shown in Fig.~\ref{fig:epochs}, the MTA of the VFL model gradually increases and then plateaus with the increase in training epochs. Simultaneously, the ASR of our attacks also shows a significant growth trend. This finding further demonstrates the alignment of goals between the attackers of our triggerless backdoor attack and the participants of the overall VFL model, i.e., better model prediction capability typically implies better backdoor attack capability.

\subsubsection{Impact of Embedding Dimensionality}

\begin{figure}[t]
    \centering
    \includegraphics[width=0.75\columnwidth]{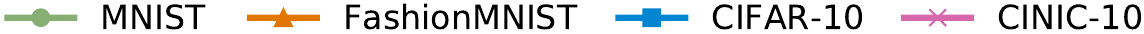}\\
    \subfloat[mASR]{\includegraphics[width=0.48\columnwidth]{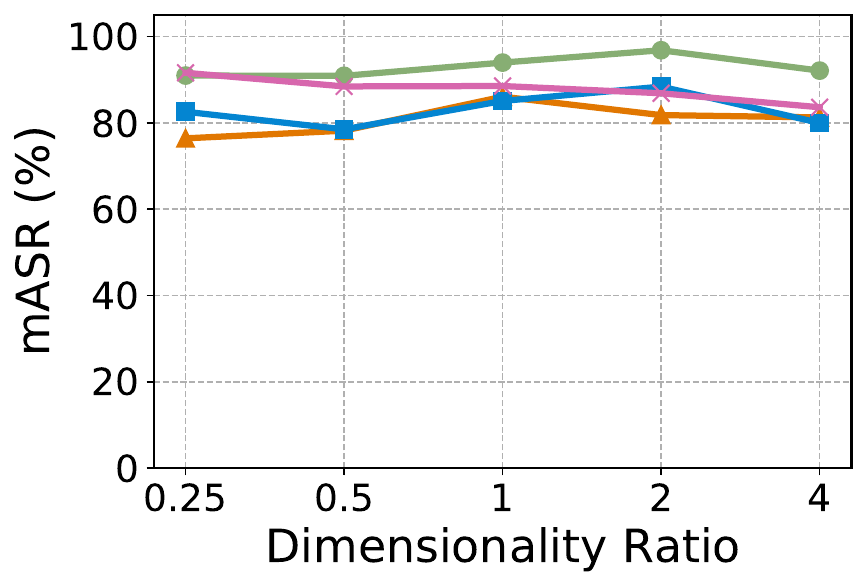}}    
    \hspace{0.02\columnwidth}
    \subfloat[rASR]{\includegraphics[width=0.48\columnwidth]{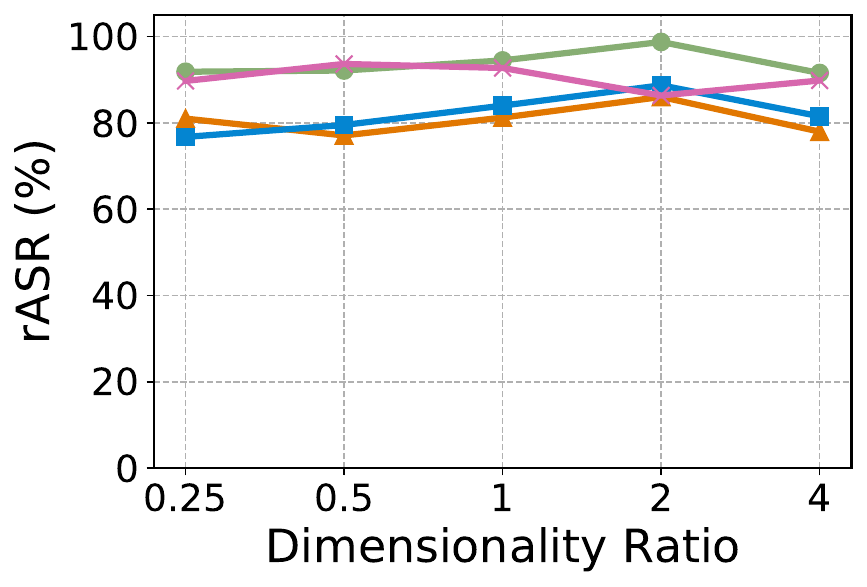}}
    \caption{Impact of embedding dimensionality on attack performance.}
    \label{fig:emb_dimension}
\end{figure}

To evaluate the impact of embedding dimensionality on our attack performance, we introduce a dimensionality ratio by scaling the embedding dimensions to 0.25, 0.5, 2, and 4 times their original values. The original embedding dimensions for each dataset are as follows: MNIST and FashionMNIST are 196, CIFAR-10 is 1024, and CINIC-10 is 512. Experimental results are presented in Fig.~\ref{fig:emb_dimension}. These results demonstrate that our attack performance is largely unaffected by changes in embedding dimensionality. We believe this is primarily because differences in embedding dimensionality do not affect the clustering of different label categories within their respective embeddings. Furthermore, under identical training settings, the main task accuracy remains largely unaffected by embedding dimensionality. Consequently, embeddings with varying dimensions exhibit comparable representational capabilities for labels, resulting in similar attack performance.

\subsubsection{Impact of Models}

The impact of the passive parties' and the attacker's model on the performance of the backdoor attack is also an important factor. We investigate this in detail in the subsequent Section~\ref{sec:emb_layer} and show the results in Tables~\ref{tab:emb_layer_all} and \ref{tab:emb_layer_single}, so we will not repeat them here.

\subsection{Performance under Defense}

In this section, we evaluate the performance of our proposed triggerless backdoor attack under nine different defense strategies from four perspectives.

\subsubsection{General Defenses}

\begin{figure*}[t]
    \centering
    \includegraphics[width=0.45\textwidth]{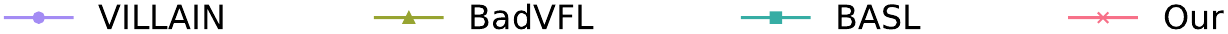}\\
	\subfloat[Clipping, MNIST]{\includegraphics[width=0.23\textwidth]{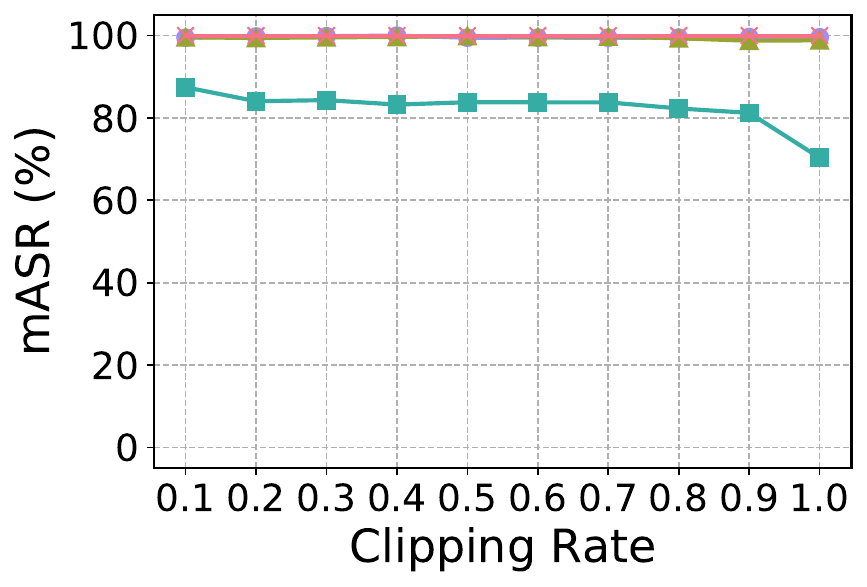}}
    \hspace{2mm}
    \subfloat[Compression, MNIST]{\includegraphics[width=0.23\textwidth]{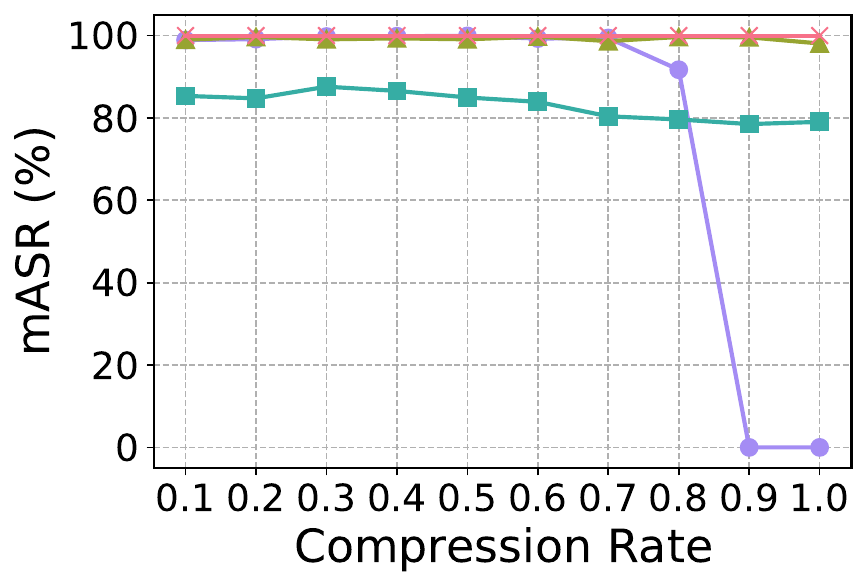}}
    \hspace{2mm}
    \subfloat[DP, MNIST]{\includegraphics[width=0.23\textwidth]{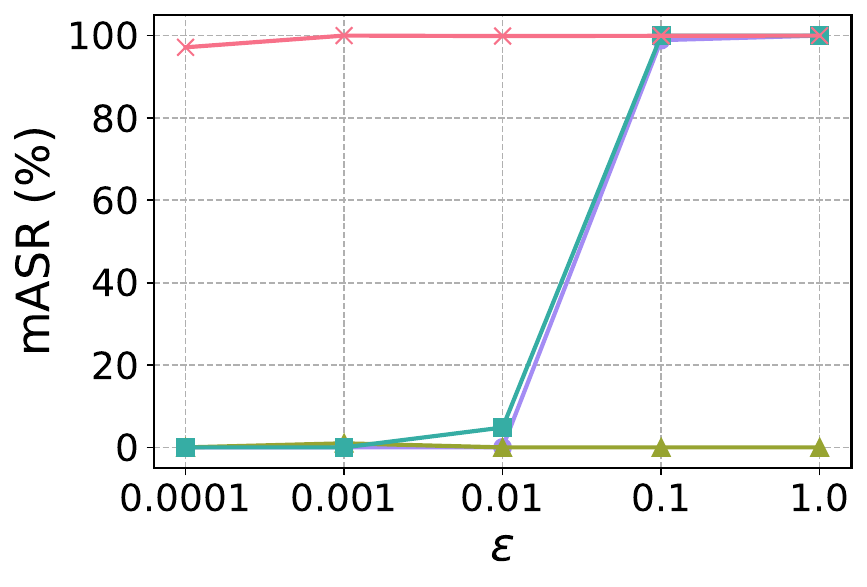}}
    \hspace{2mm}
    \subfloat[Detection, MNIST]{\includegraphics[width=0.23\textwidth]{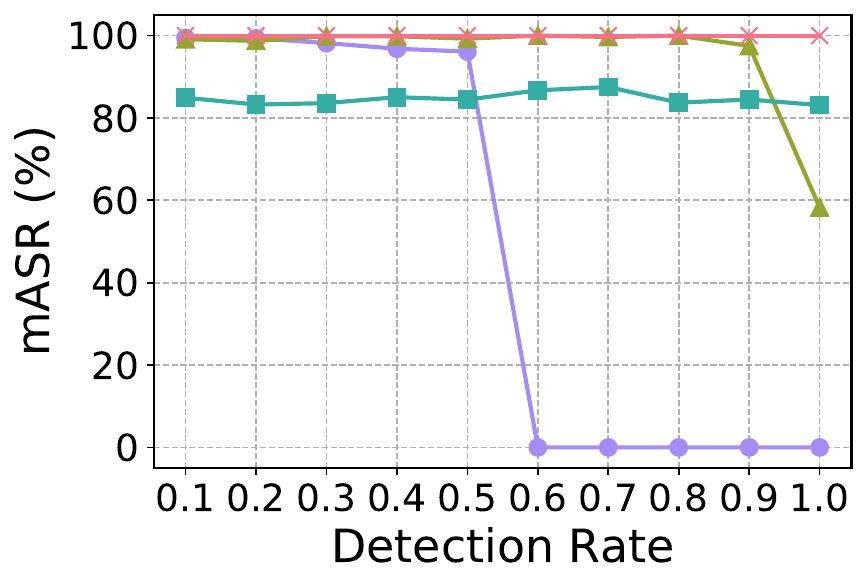}}\\
    \subfloat[Clipping, CIFAR-10]{\includegraphics[width=0.23\textwidth]{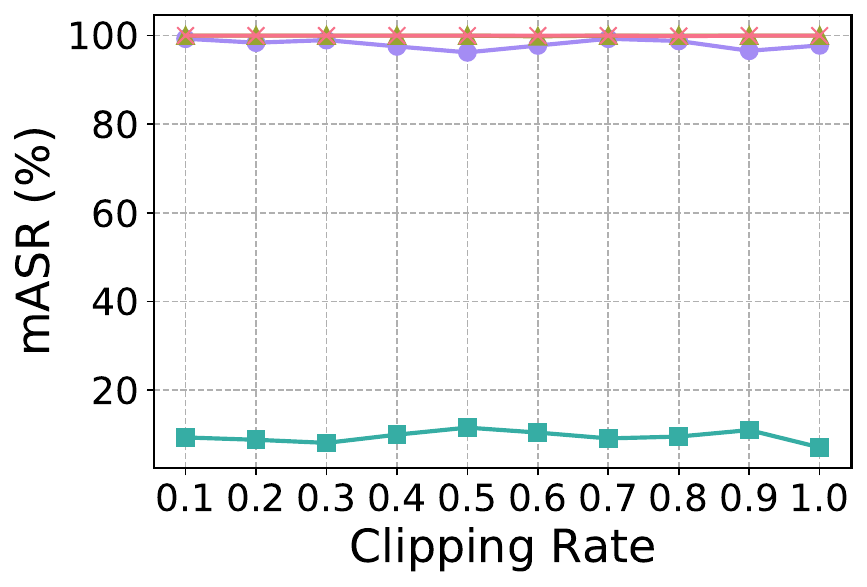}}
    \hspace{2mm}
    \subfloat[Compression, CIFAR-10]{\includegraphics[width=0.23\textwidth]{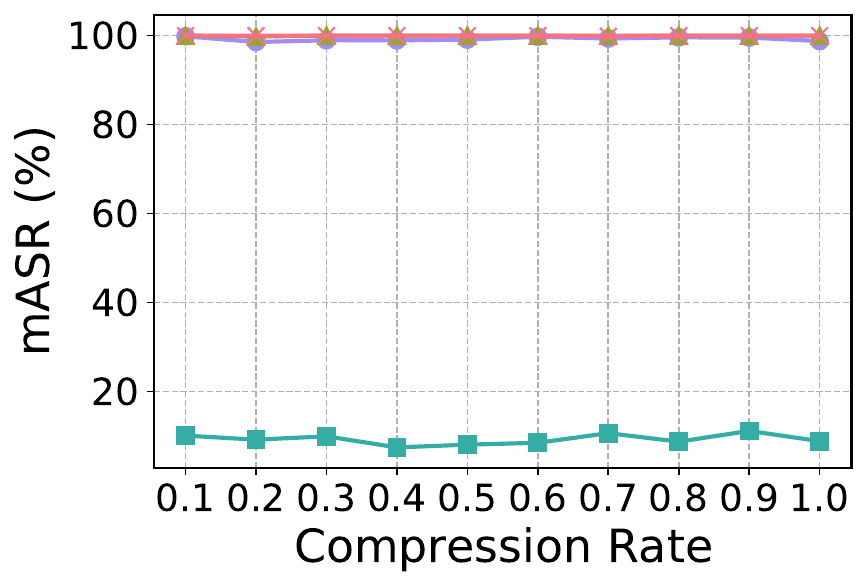}}
    \hspace{2mm}
    \subfloat[DP, CIFAR-10]{\includegraphics[width=0.23\textwidth]{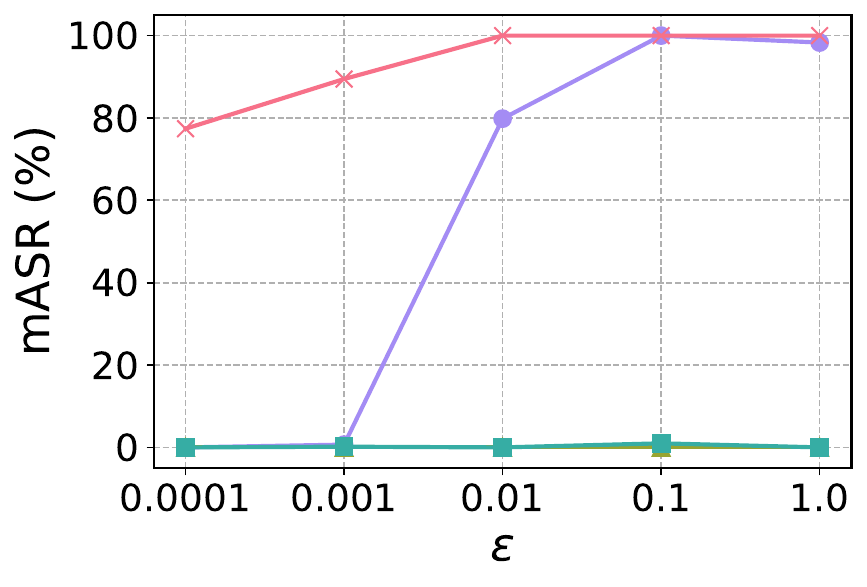}}
    \hspace{2mm}
    \subfloat[Detection, CIFAR-10]{\includegraphics[width=0.23\textwidth]{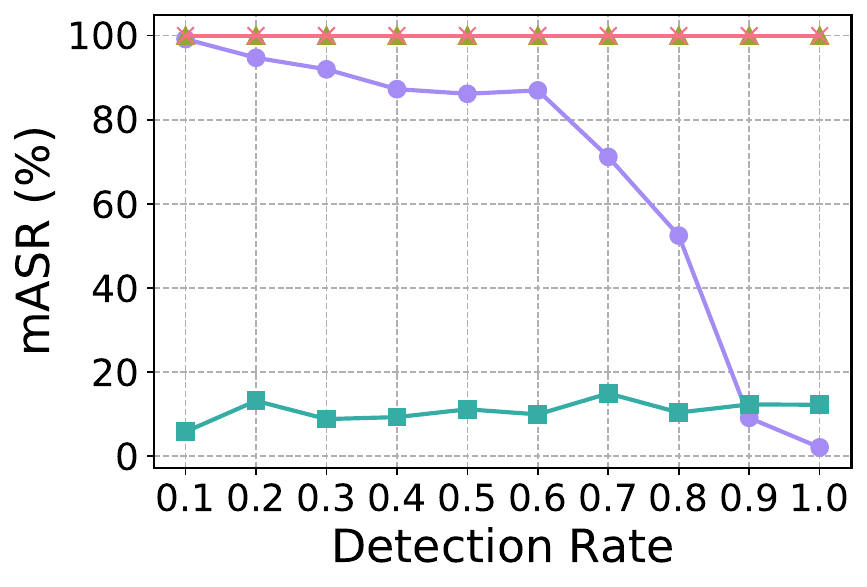}}\\
    \caption{Attack performance under different defenses with different defense rates.}
    \label{fig:defense}
\end{figure*}

We first evaluate the performance of our triggerless backdoor attack under three general defense strategies, including gradient clipping, gradient compression and differential privacy. The evaluation consists of three aspects: (\romannumeral1) the resilience and performance of our attack across various defense strategies, (\romannumeral2) a comparison between our attack and conventional baseline backdoor attacks under these defenses, and (\romannumeral3) the impact of varying defense rates on the effectiveness of each backdoor attack strategy.

To ensure a fair comparison, we conduct the experiments in a two-party VFL scenario rather than a multi-party VFL scenario. This decision was primarily based on two factors: first, some backdoor attacks are specifically designed for two-party VFLs and are not applicable to multi-party VFLs (the performance of these backdoor attacks in two-party VFLs is detailed in Appendix~B). Second, the results in Table~\ref{tab:comparison_passive4} show that these attacks exhibit a low attack success rate in multi-party VFLs, making it less meaningful to evaluate the impact of defense strategies on attacks in the context of such low success rates.

\paragraph{Gradient Clipping} Gradient clipping defends against backdoor attacks by clipping gradients that are excessively large or small~\cite{fang2023improved,mao2023differential}. This prevents attackers from pinpointing and manipulating the most influential features by implanting triggers. Our attack employs a norm-based gradient clipping strategy, where an elevated clipping rate corresponds to a higher proportion of gradients being clipped. As illustrated in Fig.~\ref{fig:defense}, the experimental results show that the ASR of the BASL exhibits a declining pattern with the increase in clipping rate, and the mASR of other baseline attacks experiences variability. Notably, our attack demonstrates resilience against this defense mechanism, maintaining robust attack efficacy despite the gradient clipping. This is because our attack is performed exclusively during the inference phase, without implanting any triggers during the training phase, thereby ensuring full compliance with the VFL protocol. Moreover, while gradient clipping restricts model updates to some extent, it cannot and should not significantly diminish the model's representation capabilities. As a result, our backdoor attack, which relies on the bottom model's ability to represent labels, is minimally affected by this defense strategy.

\paragraph{Gradient Compression} As a widely adopted defense strategy in federated learning, gradient compression aims at reducing communication overhead and enhancing privacy by minimizing the size of transmitted gradients~\cite{gu2023lr,chen2023practical,xuan2023practical}. A higher compression rate results in less data being transmitted. According to the results depicted in Fig.~\ref{fig:defense}, gradient compression proves to be an effective defense, particularly against the VILLAIN and BASL attacks, although it suggests that a robust defense may require a substantial compression rate. However, gradient compression marginally impacts the prediction accuracy of the final model, allowing attackers to exploit a well-trained bottom model. Consequently, the defense strategy of gradient compression offers limited protection against our triggerless backdoor attack.

\paragraph{Differential Privacy} Noisy gradient is a common defense strategy, where differential privacy~(DP)~\cite{dwork2006differential,abadi2016deep,miao2024efficient} is widely used because it introduces noise and quantifies privacy simultaneously. For DP, the privacy budget $\varepsilon$ acts as a privacy metric, with lower $\varepsilon$ values indicating stronger privacy preservation. The results in Fig.~\ref{fig:defense} reflect this. Additionally, we observe that while DP impacts our attack by diminishing the label representation capacity of the bottom model through gradient alteration, this limitation can be mitigated by extending the training epochs.

\subsubsection{Defenses against Label Inference}

Since all backdoor attacks in VFL scenarios involve label inference processes, adopting a defense strategy against label inference attacks may be effective in limiting the effectiveness of backdoor attacks. Therefore, we choose the current state-of-the-art defense strategy against label inference attacks to evaluate the performance of our proposed triggerless backdoor attack.

\paragraph{Marvell} Marvell is a noise perturbation-based defense~\cite{li2022label}. It prevents label leakage by adding noise to the model gradient and employs an optimizer to get the optimal noise distribution for achieving the trade-off between model performance and privacy preservation. Table~\ref{tab:defense_practical} shows the effectiveness of the Marvell defense against our attack. We find that Marvell can provide limited defense. This is primarily because Marvell's addition of noise to the gradient makes the overall VFL model more dependent on the top model, thereby reducing the bottom model's ability to characterize labels. However, this limitation on the bottom model's ability is constrained because the noise added is limited to ensure the model's predictive performance. Since our attack relies more on the performance of the final bottom model, Marvell's defense against our attack is limited.

\begin{table}[t!]
    \centering
    \caption{Attack performance under different defenses}
    \label{tab:defense_practical}
    \begin{threeparttable}
        \begin{tabular}{cccc}
            \toprule
            Dataset & Defense & mASR & rASR \\
            \midrule
            \multirow{5}{*}{MNIST} & None & $93.97\pm3.97\%$ & $94.47\pm7.17\%$ \\
            & Marvell & $80.30\pm9.33\%$ & $84.61\pm11.64\%$ \\
            & CAE & $86.04\pm6.80\%$ & $92.46\pm7.01\%$ \\
            & ANP & $89.81\pm4.28\%$ & $94.23\pm4.57\%$ \\
            & ABL & $91.52\pm6.74\%$ & $93.83\pm6.42\%$ \\
            \midrule
            \multirow{5}{*}{FashionMNIST} & None & $86.06\pm8.83\%$ & $81.22\pm11.64\%$ \\
            & Marvell & $81.49\pm9.19\%$ & $88.45\pm9.32\%$ \\
            & CAE & $78.69\pm10.68\%$ & $76.08\pm10.31\%$ \\
            & ANP & $86.72\pm7.57\%$ & $77.11\pm13.83\%$ \\
            & ABL & $84.27\pm7.71\%$ & $82.70\pm7.47\%$ \\
            \midrule
            \multirow{5}{*}{CIFAR-10} & None & $85.08\pm11.38\%$ & $83.99\pm12.81\%$ \\
            & Marvell & $77.51\pm10.85\%$ & $74.14\pm14.14\%$ \\
            & CAE & $84.54\pm11.06\%$ & $80.57\pm13.87\%$ \\
            & ANP & $84.45\pm8.26\%$ & $83.68\pm13.68\%$ \\
            & ABL & $84.13\pm13.65\%$ & $83.87\pm12.16\%$ \\
            \midrule
            \multirow{5}{*}{CINIC-10} & None & $88.54\pm13.22\%$ & $92.72\pm6.86\%$ \\
            & Marvell & $80.47\pm19.00\%$ & $86.61\pm12.38\%$ \\
            & CAE & $88.63\pm14.74\%$ & $87.67\pm15.08\%$ \\
            & ANP & $90.62\pm8.40\%$ & $89.61\pm10.51\%$ \\
            & ABL & $88.21\pm12.00\%$ & $87.69\pm13.05\%$ \\
            \bottomrule
        \end{tabular}
    \end{threeparttable}
\end{table}

\paragraph{Confusional Autoencoder~(CAE)} CAE is a defense strategy aimed at preventing label inference attacks by training VFL models with obfuscated labels~\cite{zou2024defending}. Specifically, CAE maps the original labels into the soft label space via an obfuscated autoencoder such that the labels of each class tend to be similar instead of the original one-hot type. As shown in Table~\ref{tab:defense_practical}, our experimental results indicate that CAE has little effect on defending against our attack, especially on complex datasets. We believe this is mainly because, even though obfuscated labels are used to train the VFL model, the goal of the training participants is still to obtain a well-trained model. Consequently, this training objective allows the attacker's bottom model to retain good label representation. Our triggerless backdoor attack leverages this strong characterization, maintaining high attack performance. Additionally, more complex training data typically requires a more complex bottom model, which increases its contribution to the overall VFL model and makes it less susceptible to CAE defenses.

\subsubsection{Defenses against Backdoor Attacks}

Next, we evaluate the performance of our attack under existing defense strategies against backdoor attacks.

\paragraph{Detection} Dynamic trigger detection is a defense strategy specifically for backdoor attacks~\cite{bai2023villain,guo2024universal}. It identifies trigger-implanted models by monitoring and analyzing the embeddings shared by the passive party across epochs. Fig.~\ref{fig:defense} demonstrates that it is an effective backdoor defense strategy against baseline attacks. However, since our attack bypasses trigger implantation and relies on the label characterization ability of the bottom model, this detection defense is ineffective against our triggerless backdoor attack.

\paragraph{Adversarial Neuron Pruning~(ANP)} Neurons implanted with backdoors typically behave differently from benign neurons in their reflection of data, so~\citet{wu2021adversarial} proposed a neuron pruning-based backdoor defense strategy by pruning out neurons that may contain backdoors. The experimental results in Table~\ref{tab:defense_practical} demonstrate that this defense strategy is largely ineffective in VFL scenarios. This is because the active party can only defend by pruning its own top model, which constitutes a small portion of the entire VFL model, and lacks the ability to modify the bottom models owned by the passive parties. Additionally, our triggerless backdoor attack does not generate malicious neurons during the training phase, rendering the pruning of malicious neurons ineffective.

\paragraph{Anti-Backdoor Learning~(ABL)} ABL is a defense strategy based on backdoor identification and suppression~\cite{li2021anti}. It avoids backdoor implantation by identifying the poisoned samples during the training process and using unlearning to suppress the influence of poisoned samples on the model. The experimental results in Table~\ref{tab:defense_practical} show that this defense strategy is almost ineffective when implementing our attack. This is mainly because that our backdoor attack does not implant any triggers in the training phase, making the backdoor unrecognizable. In addition, we also observe that this defense strategy against conventional backdoor attacks performs worse than the defense strategies against label inference, since our attack relies more on the label inference process.

\subsubsection{Specific Defense in the Inference Phase}

Given that existing defenses predominantly focus on the training phase while overlooking defense during inference, we propose a novel backdoor attack defense strategy specifically applied in the inference phase. This strategy is tailored to counter our proposed backdoor attack, leveraging the attacker's reliance on amplification factors to mislead the top model in multi-party VFL scenarios. The defense operates by monitoring the $L_2$ norm of embeddings: the active party records the maximum $L_2$ norm of each passive party's embeddings once the model has stabilized during training. During inference, the active party calculates the $L_2$ norm of each receiving embedding and compares it to the previously recorded maximum. If the current embedding's $L_2$ norm exceeds this threshold, it is flagged as a potential backdoor attack, and the malicious embedding is replaced to neutralize the threat; otherwise, standard inference proceeds.

\begin{table}[t!]
    \centering
    \caption{Attack performance under the specific defense in the inference phase}
    \label{tab:defense_dm}
    \begin{threeparttable}
        \begin{tabular}{ccc}
            \toprule
            Dataset & mASR & rASR \\
            \midrule
            MNIST & $0.08\pm0.06\%$ & $0.00\pm0.00\%$ \\
            FashionMNIST & $6.55\pm3.27\%$ & $6.62\pm2.96\%$ \\
            CIFAR-10 & $36.47\pm10.89\%$ & $35.06\pm11.28\%$ \\
            CINIC-10 & $62.78\pm13.12\%$ & $59.51\pm11.89\%$ \\
            \bottomrule
        \end{tabular}
    \end{threeparttable}
\end{table}

We evaluate the performance of our triggerless backdoor attack against the aforementioned specific defense across four datasets, with results presented in Table~\ref{tab:defense_dm}. We observe that for datasets with relatively simple features, such as MNIST and FashionMNIST, this defense strategy successfully mitigates our backdoor attack. However, as dataset complexity increases, the defense's effectiveness diminishes. This decline is primarily attributed to the greater diversity of embeddings generated by complex datasets, which results in a more dispersed embedding distribution. Such loosely clustered and highly personalized embeddings often exhibit larger $L_2$ norms, making it more challenging for the defense to distinguish and neutralize malicious embeddings. Considering that real-world data typically possesses high feature complexity, we emphasize the imperative for the VFL security community to develop more refined and targeted defense strategies against our triggerless backdoor attack.

\section{Discussion}

In this section, we first discuss two additional ways that may affect the performance of our proposed attack---embedding layers and learning rate adjustment. Then we discuss the limitation of our proposed attack and potential future works.

\subsection{Embedding Layer}
\label{sec:emb_layer}

Consider that our proposed triggerless backdoor attack relies on the label representation capability of the bottom model, which may be enhanced by introducing an embedding layer. Therefore, we discuss the impact of the embedding layer on our attack performance.

We first evaluate the performance of our attack when all passive parties use the same bottom model and introduce embedding layers simultaneously. The experiments are conducted on two datasets, CIFAR-10 and CINIC-10, with all the bottom models of the passive parties utilizing a convolutional neural network structure. Additionally, we introduce an additional fully connected layer after this convolutional neural network as the embedding layer for the bottom model. In the experiments, the number of fully connected layers is increased from 0 to 3. The results are shown in Table~\ref{tab:emb_layer_all}. We observe that adding embedding layers enhances the performance of our attacks, with the attack performance gradually improving as the number of embedding layers increases. This improvement is primarily because the embedding layer allows the bottom model to better characterize labels, resulting in more representative embeddings. Our attack leverages these embeddings to implement the backdoor attack, thereby improving its performance.

\begin{table}[t!]
    \centering
    \caption{The impact of introducing the embedding layer to all passive parties' bottom models}
    \label{tab:emb_layer_all}
    \begin{threeparttable}
        \begin{tabular}{cccc}
            \toprule
            Dataset & Model & mASR & rASR \\
            \midrule
            \multirow{4}{*}{CIFAR-10} & CNN-FC0 & $85.08\pm11.38\%$ & $83.99\pm12.81\%$ \\
            & CNN-FC1 & $92.34\pm8.99\%$ & $96.75\pm4.14\%$ \\
            & CNN-FC2 & $96.86\pm3.02\%$ & $98.22\pm1.78\%$ \\
            & CNN-FC3 & $98.54\pm1.12\%$ & $97.99\pm1.22\%$ \\
            \midrule
            \multirow{4}{*}{CINIC-10} & CNN-FC0 & $72.35\pm16.36\%$ & $71.57\pm18.52\%$ \\
            & CNN-FC1 & $80.38\pm8.68\%$ & $81.31\pm10.87\%$ \\
            & CNN-FC2 & $88.54\pm8.42\%$ & $89.38\pm7.52\%$ \\
            & CNN-FC3 & $91.69\pm1.62\%$ & $91.42\pm2.91\%$ \\
            \bottomrule
        \end{tabular}
    \end{threeparttable}
\end{table}

\begin{table}[t!]
    \centering
    \caption{The impact of introducing the embedding layer only to the attacker's bottom model}
    \label{tab:emb_layer_single}
    \begin{threeparttable}
        \begin{tabular}{cccc}
            \toprule
            Dataset & Model & mASR & rASR \\
            \midrule
            \multirow{4}{*}{CIFAR-10} & CNN-FC0 & $85.08\pm11.38\%$ & $83.99\pm12.81\%$ \\
            & CNN-FC1 & $99.66\pm0.49\%$ & $99.93\pm0.10\%$ \\
            & CNN-FC2 & $99.96\pm0.06\%$ & $100.00\pm0.00\%$ \\
            & CNN-FC3 & $99.92\pm0.11\%$ & $100.00\pm0.00\%$ \\
            \midrule
            \multirow{4}{*}{CINIC-10} & CNN-FC0 & $72.35\pm16.36\%$ & $71.57\pm18.52\%$ \\
            & CNN-FC1 & $97.68\pm3.29\%$ & $99.29\pm1.00\%$ \\
            & CNN-FC2 & $98.05\pm3.38\%$ & $99.76\pm0.41\%$ \\
            & CNN-FC3 & $98.35\pm2.11\%$ & $99.64\pm0.38\%$ \\
            \bottomrule
        \end{tabular}
    \end{threeparttable}
\end{table}

We then evaluate the performance impact on our attack when introducing embedding layers only to the attacker's bottom model. In this scenario, we consider a \textit{malicious} rather than an \textit{honest-but-curious} attacker during the training phase, who can enhance the performance of their backdoor attack by adding embedding layers to the original bottom model. Similar to the previous experiment, but in this case, we only add embedding layers to the attacker's bottom model, increasing the number of layers from 0 to 3. Table~\ref{tab:emb_layer_single} shows the results of this experiment. We observe that the active and malicious addition of embedding layers by the attacker positively affects the attack performance, consistent with the findings of previous experiments.

In summary, the embedding layer can enhance the performance of our triggerless backdoor attack by improving the label representation capability of the bottom model. This improvement is particularly significant when the embedding layer is added to the attacker's bottom model. Therefore, the malicious attacker can add an embedding layer to their bottom model to enhance their attack performance.

\subsection{Learning Rate Adjustment}

Learning rate adjustment is a technique used to enhance attack performance in VFL scenarios, and it has been employed in several works with good results~\cite{bagdasaryan2020how,fu2022label,bai2023villain}. The main principle is that the attacker increases the contribution of its own data to the overall model by increasing the learning rate during the training phase, thus enhancing the label representation capability of the bottom model. The attacker can then exploit this well-characterized bottom model to perform more accurate label inference attacks or backdoor attacks. This technique implies that the attacker's security assumptions during the training phase are typically malicious, consistent with existing security assumptions for backdoor attacks~\cite{bai2023villain,naseri2023badvfl,he2023backdoor}. However, our proposed triggerless backdoor attack has a more stringent security assumption that the attacker is \textit{honest-but-curious} rather than \textit{malicious} during the training phase. Therefore, when discussing the impact of learning rate adjustment on the performance of our attack, we adjust the security assumptions to be consistent with those of other backdoor attacks.

Our evaluation consists of two aspects: the impact of increasing the attacker's learning rate on the performance of the attack and the impact of decreasing the learning rate, as shown in Tables~\ref{tab:increase_lr_mASR} and \ref{tab:decrease_lr_mASR}, respectively. We conduct experiments on four datasets and use mASR as the metric, with the batch size of 128 and the training epoch of 10. The experimental results for rASR as a metric are presented in Appendix~B.

\begin{table}[t!]
    \centering
    \caption{The mASR when the attacker increase the learning rate~(lr)}
    \label{tab:increase_lr_mASR}
    \begin{threeparttable}
        \begin{tabular}{cccc}
            \toprule
            Dataset & $lr=0.01$ & $lr=0.1$ & $lr=1$ \\
            \midrule
            MN & $70.61\pm0.82\%$ & $94.79\pm3.17\%$ & $98.08\pm2.03\%$ \\
            FM & $88.62\pm12.04\%$ & $95.31\pm6.26\%$ & $98.12\pm2.66\%$ \\
            CF & $50.55\pm22.26\%$ & $78.26\pm9.48\%$ & $97.96\pm2.89\%$ \\
            CN & $47.86\pm22.84\%$ & $86.07\pm13.47\%$ & $99.77\pm0.32\%$ \\
            \bottomrule
        \end{tabular}
        \begin{tablenotes}
            \footnotesize
            \item[] The learning rate for other benign passive parties is set to 0.01.
        \end{tablenotes}
    \end{threeparttable}
\end{table}

\begin{table}[t!]
    \centering
    \caption{The mASR when the attacker decrease the learning rate~(lr)}
    \label{tab:decrease_lr_mASR}
    \begin{threeparttable}
        \begin{tabular}{cccc}
            \toprule
            Dataset & $lr=1$ & $lr=0.1$ & $lr=0.01$ \\
            \midrule
            MN & $93.97\pm3.97\%$ & $31.65\pm6.80\%$ & $2.96\pm3.05\%$ \\
            FM & $86.63\pm8.41\%$ & $18.18\pm4.62\%$ & $9.52\pm0.61\%$ \\
            CF & $85.08\pm11.38\%$ & $37.70\pm8.46\%$ & $2.51\pm3.54\%$ \\
            CN & $88.54\pm13.22\%$ & $48.87\pm12.52\%$ & $11.69\pm3.08\%$ \\
            \bottomrule
        \end{tabular}
        \begin{tablenotes}
            \footnotesize
            \item[] The learning rate for other benign passive parties is set to 1.
        \end{tablenotes}
    \end{threeparttable}
\end{table}

In the experiment of increasing the learning rate, we set the learning rate of the benign passive parties to 0.01 and gradually increase the attacker's learning rate. According to the results in Table~\ref{tab:increase_lr_mASR}, we observe that the performance of our attacks increases significantly as the attacker's learning rate increases. This indicates that increasing the learning rate can effectively improve attack performance. However, the attacker should be cautious when setting the learning rate to avoid excessive values that may lead to overfitting or detection by the active party.

In the experiment of decreasing the learning rate, we set the learning rate of the benign passive parties to 1 and gradually decrease the attacker's learning rate. As shown in Table~\ref{tab:decrease_lr_mASR}, the performance of the backdoor attack drops significantly as the learning rate decreases. The degree of this decrease is much greater than the increase in attack performance when the learning rate is raised. This indicates that the negative effect of decreasing the learning rate is much larger than the positive effect of increasing it.

Based on the above, learning rate adjustment is an effective technique for enhancing the performance of our proposed attack. The attacker can increase the learning rate to improve the attack performance, but should avoid decreasing the learning rate to ensure the best attack performance.

\subsection{Limitations}

Although our proposed triggerless backdoor attack has demonstrated strong performance across numerous experiments, like previous traditional backdoor attack methods, it still has some potential limitations due to the specific characteristics of the VFL scenario.

One limitation is that, since the passive party in VFL only possesses features but not labels, an attacker aiming to execute a targeted backdoor attack must first infer the labels of the samples. Consequently, the effectiveness of our attack is partially dependent on label inference accuracy.

Another limitation is that our backdoor attack directly replaces embeddings during the inference phase with malicious embeddings, making the attack's success rate heavily reliant on the quality of these embeddings. The construction of malicious embeddings, in turn, depends significantly on the bottom model's ability to characterize the labels derived from the attacker's feature data. While realistic VFL scenarios typically ensure that the features held by different passive parties are of comparable importance to balance the costs and benefits of training participants, extreme cases cannot be ruled out. In such scenarios, the attacker may possess the least significant features, resulting in malicious embeddings that degrade the performance of the backdoor attack.

Additionally, the number of passive parties involved in VFL training could pose another limitation. As the number of passive parties increases, malicious embeddings may be diluted by benign data from other parties. Furthermore, the introduction of the amplification factor could make the attack more easily detectable in such scenarios. However, we believe the number of passive parties has a limited impact on the performance of our attack. Unlike HFL scenarios, the number of passive parties in VFL scenarios is typically not excessively large, which helps mitigate this issue.

\subsection{Future Work}

For the triggerless backdoor attack we propose, malicious embeddings play a crucial role. Given the variations in embedding details across different VFL tasks, and to maintain a concise paper structure, we leverage statistically robust features from the embeddings to design our attack methodology and focus on detailing this novel triggerless backdoor pathway. However, the interpretability of the relationship between embeddings and labels across different VFL tasks could further drive the development of related attacks and defenses. Given the unique characteristics of embeddings in VFL scenarios, these attacks may extend beyond backdoor attacks to include label inference and feature inference attacks. Therefore, in future work, we plan to conduct more systematic and in-depth research on how embedding details influence various attacks in VFL scenarios. Specifically, we aim to investigate how the sparsity of embeddings affect attack performance while exploring methods to enhance embedding representations. Furthermore, we will analyze which dimensions primarily contribute to an embedding's ability to represent labels and strive to uncover universal patterns in this relationship.

In addition, we observe that anomaly detection defense strategies based on embedding magnitude are insufficient to counter our triggerless attacks on complex datasets. We believe these defenses still face several unresolved challenges. For example, directly identifying malicious embeddings by comparing differences between passive parties may not be reasonable: 1) passive parties may collude, making it difficult to assess their trustworthiness; 2) natural variations in feature distributions~\cite{liu2025attackers} and bottom model architectures across passive parties result in highly diverse embedding representations, leaving no consistent standard for comparison. Moreover, comparing embeddings between the training and inference phases offers limited effectiveness, as the distributions of training and test sets may differ, and certain highly distinctive samples may naturally have representations outside the expected range. Therefore, efficiently identifying malicious embeddings without affecting benign ones remains a significant challenge. In summary, our future work will focus on developing defense strategies capable of effectively countering backdoor attacks during the inference phase, particularly in complex real-world VFL scenarios.

\section{Conclusion}

In the paper, we have disclosed a new triggerless backdoor attack pathway in VFL scenarios. Unlike existing backdoor attacks that implant triggers during the training phase, our attack bypasses trigger implantation and directly misleads the VFL model during the inference phase by replacing embeddings with malicious ones. We extensively evaluate the performance of our proposed attack on five datasets under various settings, demonstrating its effectiveness and robustness. Furthermore, we have assessed the performance of our attack against nine different defense strategies, including our proposed specific defense during the inference phase. The results indicate that existing defenses are largely ineffective against our triggerless backdoor attack. We hope our work will raise awareness in the VFL security community about this new type of backdoor attack and inspire further research into effective defense strategies.

\section*{Acknowledgments}
This work was supported by the National Natural Science Foundation of China under Grant 62572007.

\small
\bibliographystyle{IEEEtranN}
\bibliography{references}

%
\appendices

\section{Algorithm}

As shown in Algorithm~\ref{alg:triggerless}, our backdoor attack contains three key modules: label inference, poison generation, and backdoor execution. The first two modules prepare the backdoor poison necessary for the attack, while the third module executes the attack. During the training phase, the attacker simply records the embeddings generated by the bottom model at each epoch, without engaging in any actions that breach the VFL protocol. During the inference phase, the attacker first infers the sample labels and records the embeddings corresponding to different labels using the label inference module. Next, the recorded embeddings are utilized to generate a valid set of malicious embeddings in the poison generation module. Finally, the attacker identifies the target samples and replaces their embeddings with the malicious embeddings to execute the backdoor attack.

\begin{algorithm}[ht]
    \caption{Feature-Based Triggerless Backdoor Attack}
    \label{alg:triggerless}
    \begin{algorithmic}[1] 
        \REQUIRE The number of labels $|l|$, the number of samples $N$, target label $l_t$, source label $l_s$, features, and labels.
        \STATE // Training Phase:
        \FOR{each epoch}
            \STATE Forward propagation and backpropagation.
            \IF{is last epoch for attacker}
                \STATE Record the embeddings $\left\{\mathrm{emb}_i^a\right\}_{i=1}^N$.
            \ENDIF
        \ENDFOR
        \STATE // Inference Phase:
        \FOR{each passive party}
            \STATE Forward propagation.
            \IF{is attacker}
                \STATE // Module \ding{202}: Label Inference
                \STATE Calculate the mean embeddings $\left\{\overline{\mathrm{aux}}_i\right\}_{i=1}^{|l|}$ for the auxiliary data associated with each label $l_i$.
                \STATE $\left\{\hat{y}_i\right\}_{i=1}^N=\mathrm{Infer}\left(\left\{\mathrm{emb}_i^a\right\}_{i=1}^N,\left\{\overline{\mathrm{aux}}_i\right\}_{i=1}^{|l|},|l|\right)$.
                \STATE Calculate the density centers $\overline{\mathrm{Emb}}=\left\{\overline{\mathrm{emb}}_i\right\}_{i=1}^{|l|}$ for different label categories by $\left\{\hat{y}_i\right\}_{i=1}^N$.
                \STATE // Module \ding{203}: Poison Generation
                \STATE Amplify $\widehat{\mathrm{emb}}=\eta\cdot\overline{\mathrm{emb}}_t$, where $\overline{\mathrm{emb}}_t$ is the density center of the target label $l_t$.
                \STATE Generate $\widetilde{\mathrm{Emb}}$, where each $\widetilde{\mathrm{emb}}_i\in\widetilde{\mathrm{Emb}}$ satisfies $\widetilde{\mathrm{emb}}_i=\widehat{\mathrm{emb}}+\delta,\ \delta\sim\mathcal{N}\left(\mu,\sigma^2\right)$.
                \STATE // Module \ding{204}: Backdoor Execution
                \FOR{each sample $i$}
                    \STATE Infer label $\hat{y}^a_i$ based on the similarity between $\overline{\mathrm{Emb}}$ and $\mathrm{emb}_i^a$.
                    \IF{$\hat{y}_i^a$ equals $l_s$}
                        \STATE Replace $\mathrm{emb}_i^a$ with $\widetilde{\mathrm{emb}}_i\in\widetilde{\mathrm{Emb}}$.
                    \ENDIF
                \ENDFOR
            \ENDIF
            \STATE Send embeddings to the active party.
        \ENDFOR
        \STATE Active party forward propagation and prediction.
    \end{algorithmic}
\end{algorithm}

\section{Additional Evaluation Results}

Table~\ref{tab:comparison_passive1} compares the performance of our proposed triggerless backdoor attack with existing backdoor attacks using triggers in a two-party VFL scenario. The results show that traditional backdoor attacks can be effectively implemented in two-party VFL. However, when combined with results from multi-party VFL scenarios, it indicates that traditional methods do not cope well with multi-party VFL. In contrast, our proposed triggerless backdoor attack is robust and effective in both two-party and multi-party VFL scenarios.

\begin{figure}[t!]
    \centering
    \subfloat[mASR]{\includegraphics[width=0.45\columnwidth]{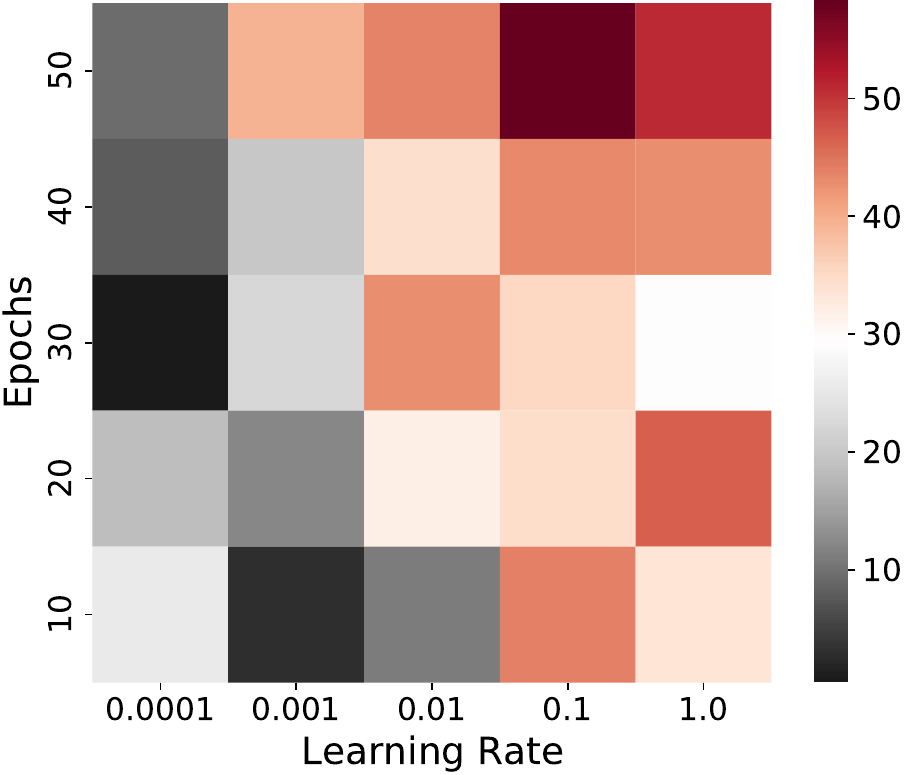}}
    \hspace{0.05\columnwidth}
    \subfloat[rASR]{\includegraphics[width=0.45\columnwidth]{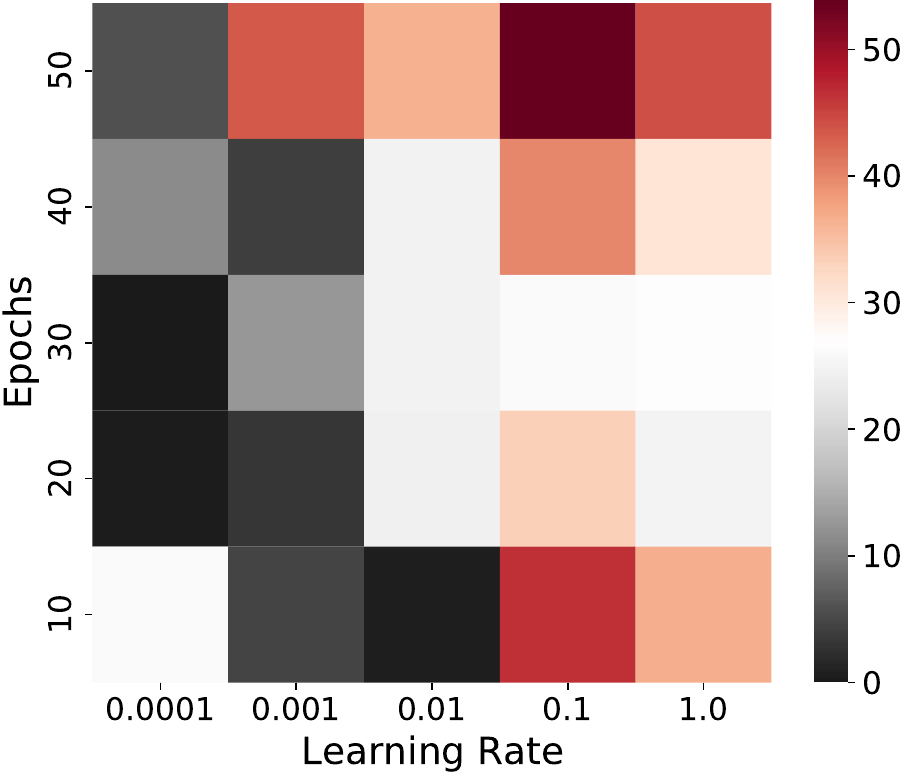}}\\
    \subfloat[mASR]{\includegraphics[width=0.45\columnwidth]{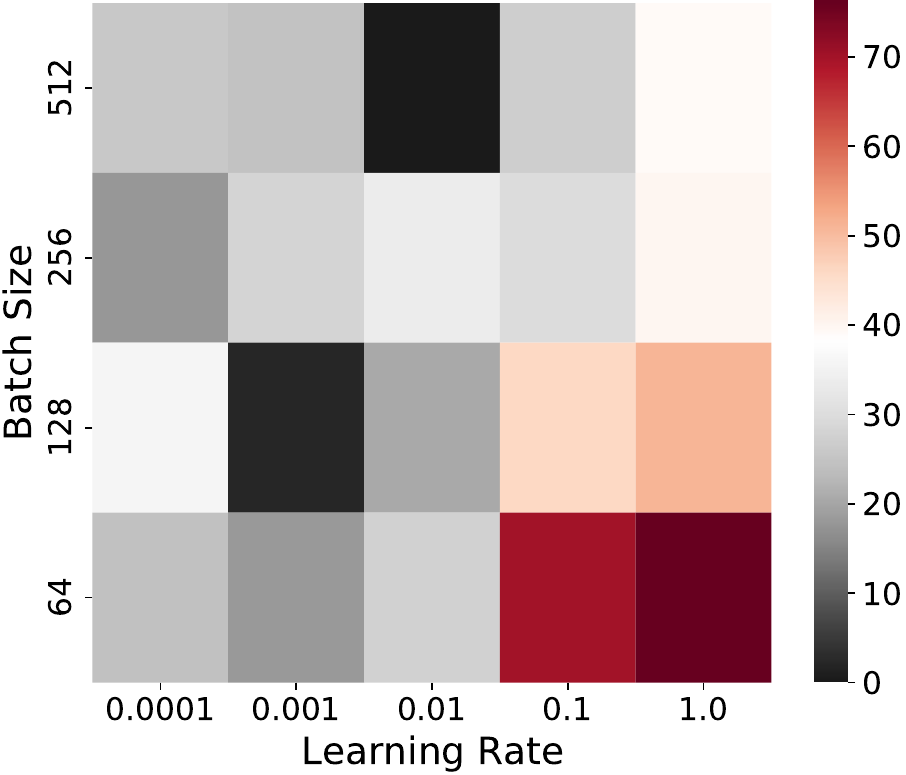}}
    \hspace{0.05\columnwidth}
    \subfloat[rASR]{\includegraphics[width=0.45\columnwidth]{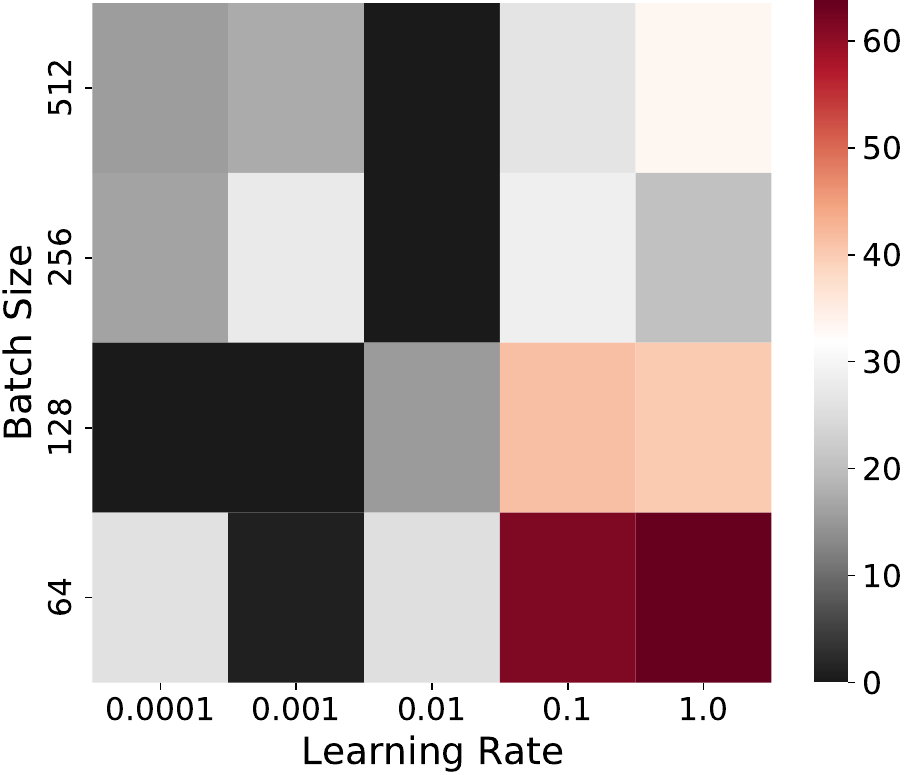}}\\
    \subfloat[mASR]{\includegraphics[width=0.45\columnwidth]{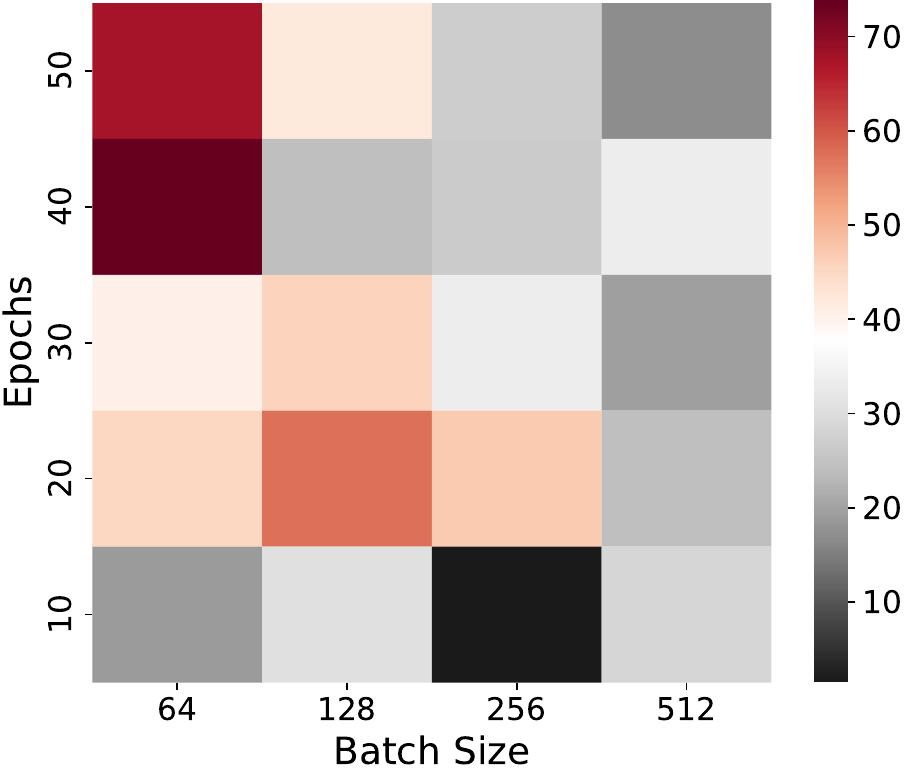}}
    \hspace{0.05\columnwidth}
    \subfloat[rASR]{\includegraphics[width=0.45\columnwidth]{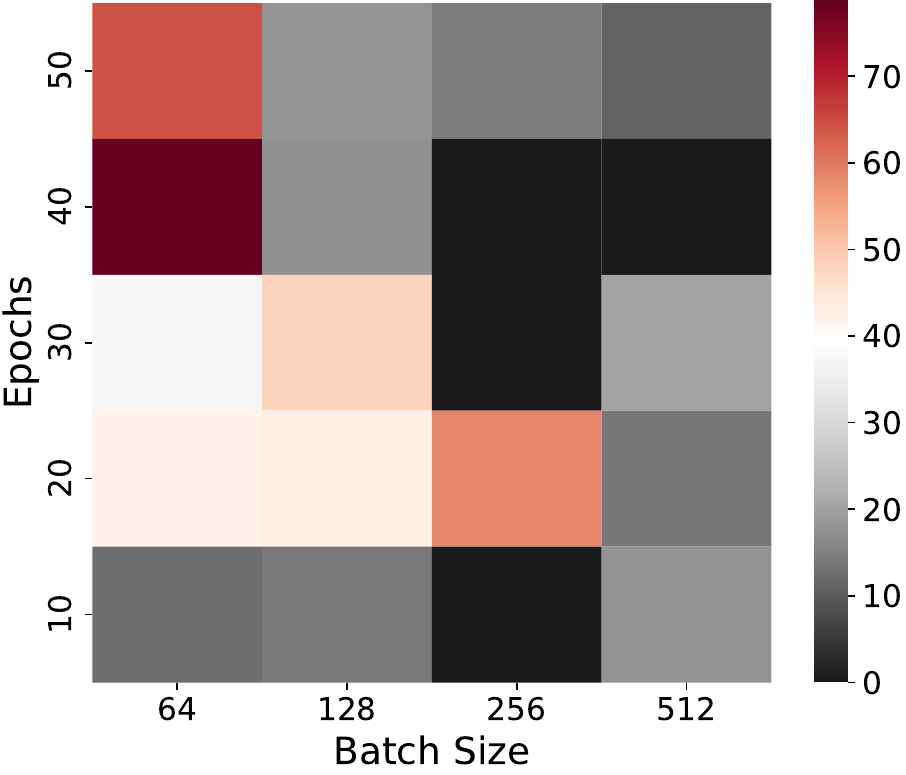}}
    \caption{Combined impact of hyperparameters on CIFAR-10 dataset.}
    \label{fig:combined}
\end{figure}

\begin{table*}[t!]
    \centering
    \caption{Attack performance of our attack with baselines in two-party VFL scenario}
    \label{tab:comparison_passive1}
    \setlength{\tabcolsep}{3.6pt}
    \begin{threeparttable}
        \begin{tabular}{ccccccccc}
            \toprule
            \multirow{2}{*}[-0.4em]{Dataset} & \multirow{2}{*}[-0.4em]{Method} & \multicolumn{3}{c}{BIR} & \multirow{2}{*}[-0.4em]{LISR} & \multicolumn{2}{c}{ASR} & \multirow{2}{*}[-0.4em]{MTA} \\
            \cmidrule{3-5}\cmidrule{7-8}
            & & tBIR & mBIR & rBIR & & mASR & rASR &  \\
            \midrule
            \multirow{4}{*}[-0.4em]{MNIST} & VILLAIN & $96.61\pm0.12\%$ & $99.73\pm0.07\%$ & $99.72\pm0.08\%$ & $73.11\pm0.52\%$ & $98.66\pm0.36\%$ & $100.00\pm0.00\%$ & $91.77\pm0.46\%$ \\
            & BadVFL & $69.84\pm27.76\%$ & $70.21\pm30.53\%$ & $70.35\pm30.42\%$ & $71.36\pm0.62\%$ & $94.93\pm2.41\%$ & $97.44\pm2.18\%$ & $77.36\pm0.38\%$ \\
            & BASL & $96.78\pm0.12\%$ & $98.44\pm0.31\%$ & $99.77\pm0.04\%$ & $72.11\pm0.43\%$ & $70.15\pm1.47\%$ & $76.31\pm2.23\%$ & $92.96\pm0.12\%$ \\
            & Our & / & / & / & $73.23\pm0.62\%$ & $\pmb{100.00\pm0.00\%}$ & $\pmb{100.00\pm0.00\%}$ & $\pmb{93.92\pm0.09\%}$ \\
            \midrule
            \multirow{4}{*}[-0.4em]{\shortstack{Fashion\\MNIST}} & VILLAIN & $89.11\pm0.05\%$ & $99.93\pm0.02\%$ & $99.96\pm0.02\%$ & $59.66\pm0.41\%$ & $100.00\pm0.00\%$ & $100.00\pm0.00\%$ & $83.80\pm0.14\%$ \\
            & BadVFL & $88.27\pm0.23\%$ & $99.32\pm0.03\%$ & $99.34\pm0.03\%$ & $58.38\pm0.32\%$ & $100.00\pm0.00\%$ & $100.00\pm0.00\%$ & $74.83\pm0.04\%$ \\
            & BASL & $86.43\pm1.80\%$ & $89.93\pm5.39\%$ & $96.99\pm1.83\%$ & $58.68\pm1.01\%$ & $62.19\pm2.41\%$ & $62.35\pm2.49\%$ & $83.56\pm0.77\%$ \\
            & Our & / & / & / & $59.40\pm0.76\%$ & $\pmb{100.00\pm0.00\%}$ & $\pmb{100.00\pm0.00\%}$ & $\pmb{84.12\pm0.09\%}$ \\
            \midrule
            \multirow{4}{*}[-0.4em]{CIFAR-10} & VILLAIN & $84.34\pm0.19\%$ & $99.17\pm0.04\%$ & $99.31\pm0.02\%$ & $21.64\pm0.63\%$ & $99.35\pm0.10\%$ & $99.73\pm0.07\%$ & $75.18\pm0.66\%$ \\
            & BadVFL & $82.04\pm0.21\%$ & $97.87\pm0.36\%$ & $98.13\pm0.34\%$ & $20.97\pm0.28\%$ & $100.00\pm0.00\%$ & $100.00\pm0.00\%$ & $50.02\pm7.42\%$ \\
            & BASL & $83.33\pm0.58\%$ & $61.73\pm4.00\%$ & $93.79\pm1.45\%$ & $21.59\pm0.20\%$ & $15.84\pm1.54\%$ & $14.20\pm1.32\%$ & $75.31\pm0.54\%$ \\
            & Our & / & / & / & $21.53\pm0.45\%$ & $\pmb{100.00\pm0.00\%}$ & $\pmb{100.00\pm0.00\%}$ & $\pmb{75.43\pm0.51\%}$ \\
            \midrule
            \multirow{4}{*}[-0.4em]{CINIC-10} & VILLAIN & $81.59\pm1.02\%$ & $99.44\pm0.03\%$ & $99.46\pm0.09\%$ & $17.82\pm0.54\%$ & $99.42\pm0.22\%$ & $99.75\pm0.14\%$ & $\pmb{61.99\pm1.36\%}$ \\
            & BadVFL & $79.06\pm0.64\%$ & $99.55\pm0.19\%$ & $99.55\pm0.22\%$ & $17.28\pm0.42\%$ & $98.11\pm2.21\%$ & $99.47\pm0.17\%$ & $38.19\pm1.69\%$ \\
            & BASL & $77.79\pm0.42\%$ & $54.11\pm5.66\%$ & $91.52\pm1.56\%$ & $17.66\pm0.52\%$ & $9.54\pm2.21\%$ & $6.94\pm2.97\%$ & $61.62\pm0.38\%$ \\
            & Our & / & / & / & $17.36\pm0.33\%$ & $\pmb{100.00\pm0.00\%}$ & $\pmb{100.00\pm0.00\%}$ & $60.97\pm2.77\%$ \\
            \midrule
            \multirow{4}{*}[-0.4em]{Criteo} & VILLAIN & $99.39\pm0.05\%$ & $99.81\pm0.04\%$ & $99.80\pm0.04\%$ & $64.47\pm1.15\%$ & $99.97\pm0.04\%$ & $99.96\pm0.05\%$ & $77.46\pm0.02\%$ \\
            & BadVFL & $99.41\pm0.10\%$ & $100.00\pm0.00\%$ & $100.00\pm0.00\%$ & $67.55\pm2.01\%$ & $100.00\pm0.00\%$ & $100.00\pm0.00\%$ & $\pmb{77.55\pm0.00\%}$ \\
            & BASL & $99.56\pm0.13\%$ & $99.56\pm0.12\%$ & $99.63\pm0.10\%$ & $66.22\pm3.84\%$ & $100.00\pm0.00\%$ & $100.00\pm0.00\%$ & $77.51\pm0.05\%$ \\
            & Our & / & / & / & $65.06\pm2.33\%$ & $\pmb{100.00\pm0.00\%}$ & $\pmb{100.00\pm0.00\%}$ & $77.54\pm0.01\%$ \\
            \bottomrule
        \end{tabular}
        \begin{tablenotes}
            \footnotesize
            \item[] The \textbf{bold} data indicate the best performance.
        \end{tablenotes}
    \end{threeparttable}
\end{table*}

Fig.~\ref{fig:combined} shows the combined impact of the learning rate, batch size, and training epochs on the attack performance, evaluated using the CIFAR-10 dataset. We observe that the learning rate and training epochs have a significant impact on the attack performance, while the batch size has a limited effect. The results of the interaction between learning rate and training epochs indicate that VFL models with higher learning rates and more training epochs typically yield better attack performance in our triggerless attacks. Additionally, experiments on the combined impact of batch size and the other two hyperparameters show that using a smaller batch size is more favorable for improving attack performance. Meanwhile, the batch size setting has little effect on the impact of the other two hyperparameters on the attack, i.e., a higher learning rate and more training epochs generally imply a higher ASR.

Tables~\ref{tab:increase_lr_rASR} and \ref{tab:decrease_lr_rASR} present the performance of our attack when the attacker uses the learning rate adjustment technique and employs rASR as a metric. The results indicate the same findings as those shown by using mASR as a metric.

\begin{table}[t!]
    \centering
    \caption{The rASR when the attacker increase the learning rate~(lr)}
    \label{tab:increase_lr_rASR}
    \begin{threeparttable}
        \begin{tabular}{cccc}
            \toprule
            Dataset & $lr=0.01$ & $lr=0.1$ & $lr=1$ \\
            \midrule
            MN & $63.62\pm11.18\%$ & $96.36\pm4.71\%$ & $98.02\pm2.70\%$ \\
            FM & $84.63\pm17.81\%$ & $98.51\pm2.11\%$ & $99.60\pm0.56\%$ \\
            CF & $45.02\pm25.28\%$ & $77.73\pm5.10\%$ & $98.21\pm2.53\%$ \\
            CN & $51.26\pm34.98\%$ & $87.10\pm14.40\%$ & $99.71\pm0.42\%$ \\
            \bottomrule
        \end{tabular}
        \begin{tablenotes}
            \footnotesize
            \item[] The learning rate for other benign passive parties is set to 0.01.
        \end{tablenotes}
    \end{threeparttable}
\end{table}

\begin{table}[t!]
    \centering
    \caption{The rASR when the attacker decrease the learning rate~(lr)}
    \label{tab:decrease_lr_rASR}
    \begin{threeparttable}
        \begin{tabular}{cccc}
            \toprule
            Dataset & $lr=1$ & $lr=0.1$ & $lr=0.01$ \\
            \midrule
            MN & $94.47\pm7.17\%$ & $33.04\pm7.54\%$ & $0.00\pm0.00\%$ \\
            FM & $84.12\pm9.64\%$ & $17.79\pm7.26\%$ & $6.56\pm4.34\%$ \\
            CF & $83.99\pm12.81\%$ & $35.16\pm5.23\%$ & $1.55\pm2.19\%$ \\
            CN & $92.72\pm6.86\%$ & $44.94\pm13.97\%$ & $6.61\pm4.05\%$ \\
            \bottomrule
        \end{tabular}
        \begin{tablenotes}
            \footnotesize
            \item[] The learning rate for other benign passive parties is set to 1.
        \end{tablenotes}
    \end{threeparttable}
\end{table}

\begin{figure}[t!]
    \centering
    \subfloat[MNIST]{\includegraphics[width=0.48\columnwidth]{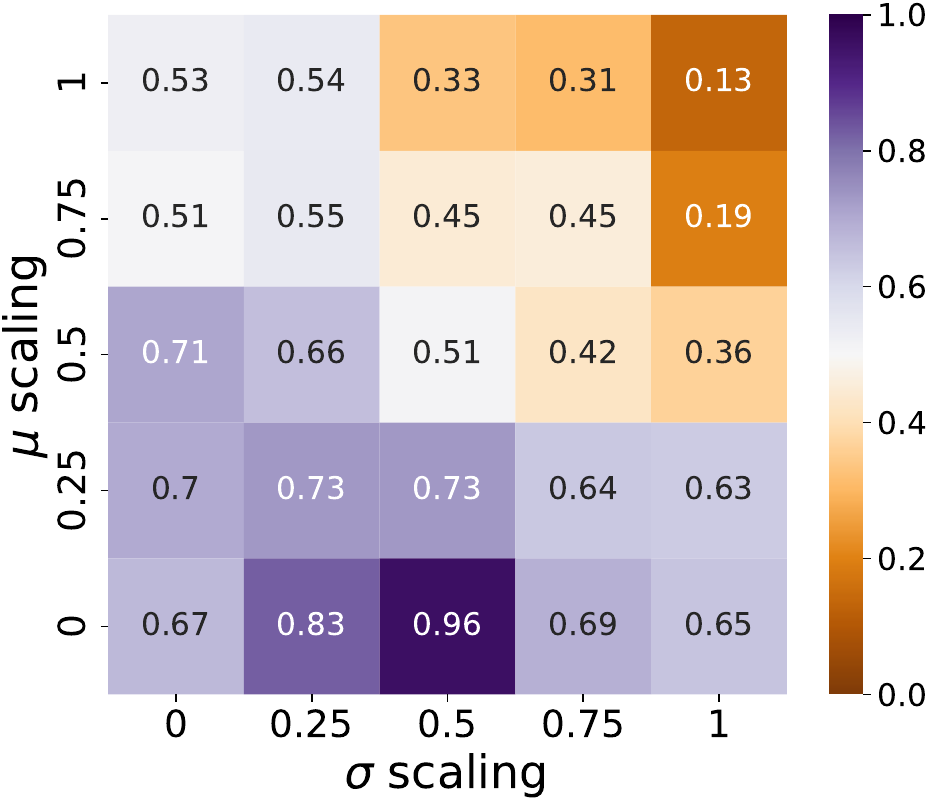}}    
    \hspace{2mm}
    \subfloat[CIFAR-10]{\includegraphics[width=0.48\columnwidth]{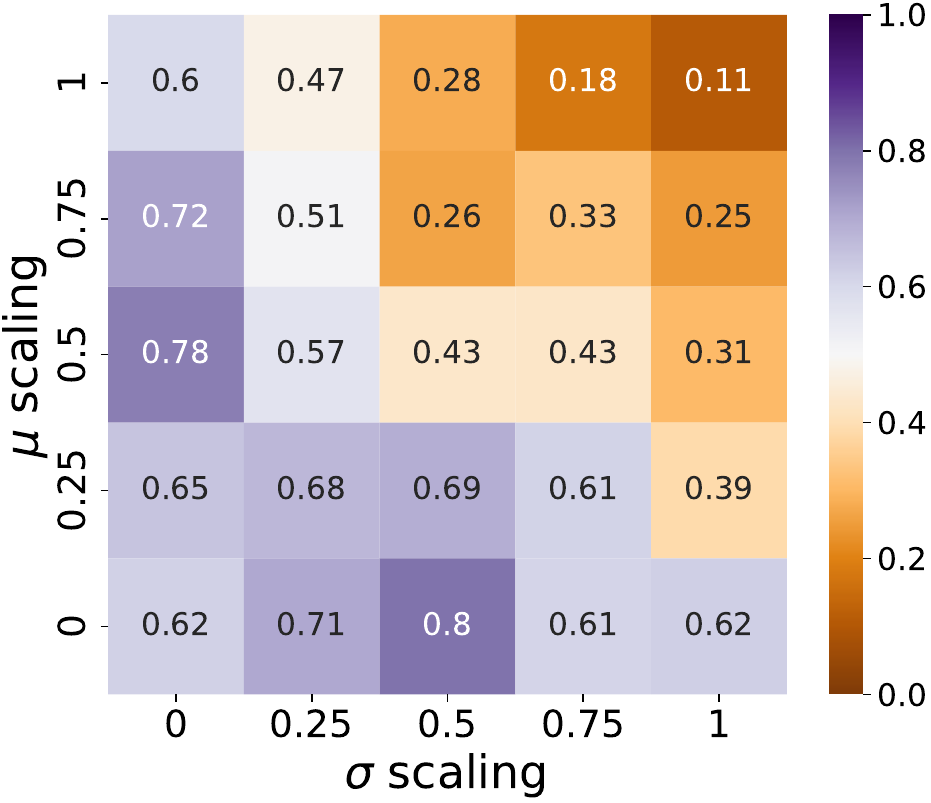}}
    \caption{Impact of the Gaussian distribution on the ASR.}
    \label{fig:gaussian_rASR}
\end{figure}

Fig.~\ref{fig:gaussian_rASR} illustrates the impact of Gaussian perturbation distribution on the performance of our attack when rASR is used as the metric. The experimental results align with those obtained using mASR as the metric.

Table~\ref{tab:overhead} summarizes the computational overhead in seconds for our proposed attack compared to traditional attacks across the test sets of five datasets. To ensure fairness and eliminate the influence of GPU and CPU arithmetic, all experiments are conducted on a four-core Intel(R) Core(TM) i7-7700 CPU. The results demonstrate that our attack requires computational resources comparable to those of other attacks, without introducing additional overhead.

\begin{table}[t!]
    \centering
    \caption{The computational overhead of different attacks}
    \label{tab:overhead}
    \begin{threeparttable}
        \begin{tabular}{ccccc}
            \toprule
            DS & VILLAIN & BadVFL & BASL & Our \\
            \midrule
            MN & $0.42\pm0.01$ & $0.46\pm0.00$ & $0.42\pm0.00$ & $0.43\pm0.01$ \\
            FM & $0.42\pm0.00$ & $0.45\pm0.02$ & $0.42\pm0.00$ & $0.44\pm0.00$ \\
            CF & $3.14\pm0.01$ & $3.67\pm0.03$ & $3.05\pm0.00$ & $3.03\pm0.00$ \\
            CN & $28.30\pm0.33$ & $37.94\pm1.29$ & $28.70\pm0.07$ & $28.34\pm0.01$ \\
            CT & $0.48\pm0.02$ & $0.62\pm0.02$ & $0.51\pm0.05$ & $0.54\pm0.05$ \\
            \bottomrule
        \end{tabular}
        \begin{tablenotes}
            \footnotesize
            \item[] We use runtime as a metric in seconds. DS denotes the dataset.
        \end{tablenotes}
    \end{threeparttable}
\end{table}

\end{document}